%% file: main.tex
\documentclass[10pt,twocolumn,letterpaper]{article}

\usepackage{wacv}  


\definecolor{wacvblue}{rgb}{0.21,0.49,0.74}
\usepackage[pagebackref,breaklinks,colorlinks,allcolors=wacvblue]{hyperref}
\usepackage{amsmath}
\usepackage{amssymb}
\usepackage{graphicx}
\usepackage{booktabs}
\usepackage{multirow}
\usepackage{caption}
\usepackage{subcaption}

\usepackage{algorithm}
\usepackage{algpseudocode}

\def\wacvPaperID{*****}
\def\confName{WACV}
\def\confYear{2026}

\title{FIRE-VLM: A Vision-Language-Driven Reinforcement Learning Framework for UAV Wildfire Tracking in a Physics-Grounded Fire Digital Twin
}

\author{
Chris Webb\\ Clemson University
\and
Mobin Habibpour\\ Clemson University
\and
Mayamin Hamid Raha\\ University of Nevada, Reno
\and
Ali Reza Tavakkoli\\ University of Nevada, Reno
\and
Janice Coen\\ NSF NCAR
\and
Fatemeh Afghah\\ Clemson University
}

\begin{document}
\maketitle


\begin{abstract}
Wildfire monitoring demands autonomous systems capable of reasoning under extreme visual degradation, rapidly evolving physical dynamics, and scarce real-world training data. Existing UAV navigation approaches rely on simplified simulators and supervised perception pipelines, and lack embodied agents interacting with physically realistic fire environments. We introduce FIRE-VLM, the first end-to-end vision-language model (VLM) guided reinforcement learning (RL) framework trained entirely within a high-fidelity, physics-grounded wildfire digital twin. Built from USGS Digital Elevation Model (DEM) terrain, LANDFIRE fuel inventories, and semi-physical fire-spread solvers, this twin captures terrain-induced runs, wind-driven acceleration, smoke plume occlusion, and dynamic fuel consumption. Within this environment, a PPO agent with dual-view UAV sensing is guided by a CLIP-style VLM. Wildfire-specific semantic alignment scores, derived from a single prompt describing active fire and smoke plumes, are integrated as potential-based reward shaping signals. Our contributions are: (1) a GIS-to-simulation pipeline for constructing wildfire digital twins; (2) a VLM-guided RL agent for UAV firefront tracking; and (3) a wildfire-aware reward design that combines physical terms with VLM semantics. Across five digital-twin evaluation tasks, our VLM-guided policy reduces time-to-detection by up to $6\times$, increases time-in-FOV, and is, to our knowledge, the first RL-based UAV wildfire monitoring system demonstrated in kilometer-scale, physics-grounded digital-twin fires.
\end{abstract}


\begin{figure*}[t]
\centering
\includegraphics[width=.8\linewidth]{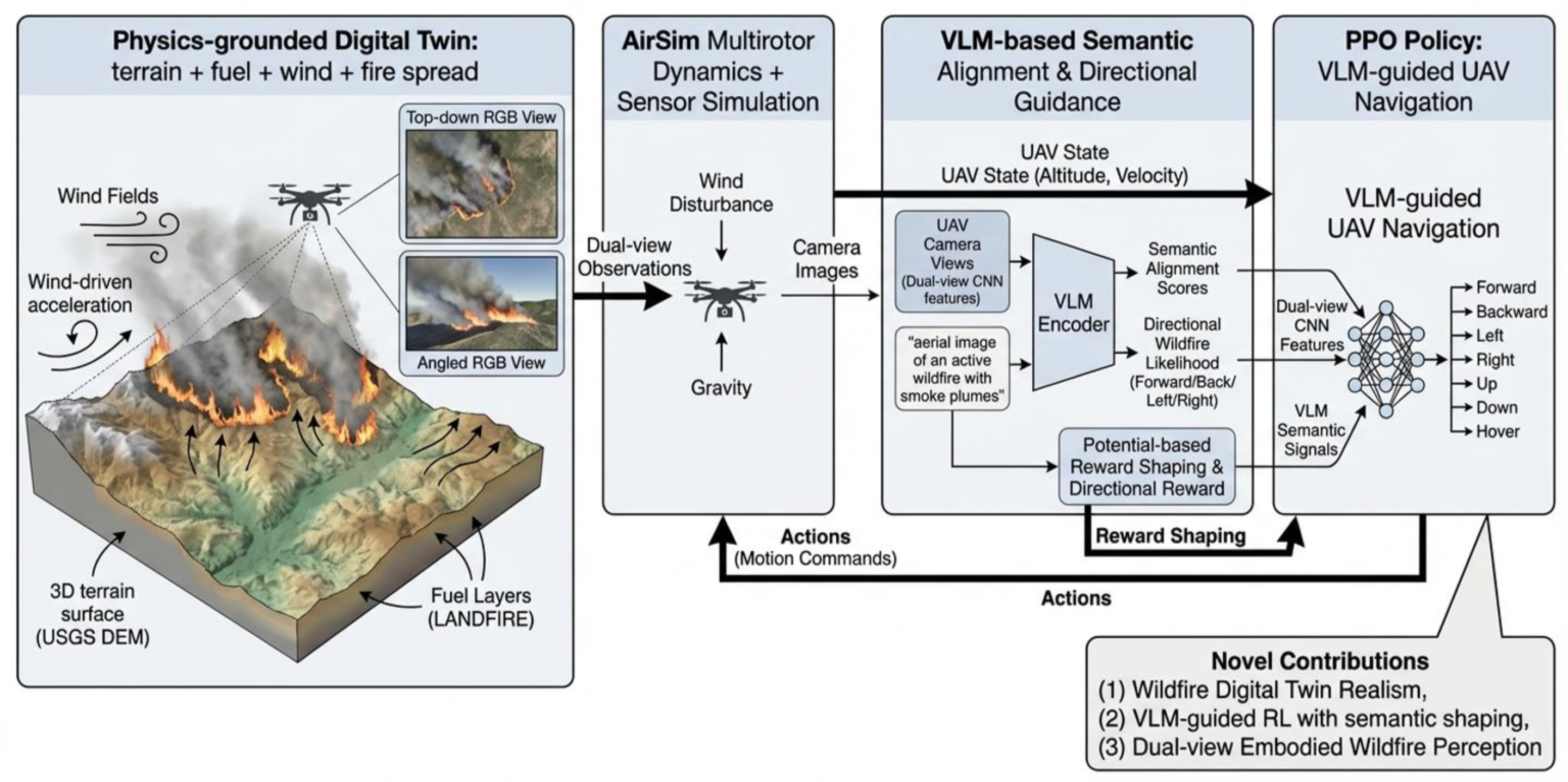}
\caption{System overview of the proposed VLM-guided UAV wildfire monitoring framework. A physics-grounded wildfire digital twin (terrain, fuel, wind, and fire spread) is simulated in AirSim, producing dual-view RGB observations from top-down and angled cameras. A pretrained VLM computes semantic alignment and directional wildfire likelihood, which shape the PPO policy’s reward and guide the UAV’s 3D motion commands for robust firefront tracking.}

\label{fig:system_overview}
\end{figure*}

\section{Introduction}

Wildfires are becoming larger, faster-moving, and more destructive, creating an urgent need for rapid, high-resolution situational awareness \cite{jolly2015climate,abatzoglou2016impact,doerr2016global}. Existing observation methods lack the timeliness and safety required, leaving incident commanders without the real-time intelligence needed for effective early-stage suppression, resource allocation, and evacuation decisions. Unmanned aerial vehicles (UAVs) provide a rapidly deployable sensing platform capable of low-altitude operation through smoke-obscured airspace while delivering high-frequency RGB/thermal observations \cite{merino2012uas}. Advances in onboard autonomy and BVLOS navigation have enabled UAVs to function in GPS-degraded, hazard-dense wildfire environments, motivating increasing interest in fully autonomous, AI-driven wildfire surveillance systems \cite{sayed2024}. However, several fundamental challenges remain: (i) \textit{Highly dynamic fire behavior:} fire fronts evolve based on wind, slope, fuel moisture, and atmospheric coupling, often changing direction rapidly; (ii) \textit{Severe visual degradation:} smoke obscures flame edges and attenuates RGB imagery, creating partial observability; (iii) \textit{Chaotic wind fields:} convective updrafts and turbulence disrupt UAV stability and complicate trajectory planning; and (iv) \textit{Limited real-world training data:} collecting labeled UAV wildfire data is dangerous, expensive, and logistically constrained. Together, these constraints make autonomous wildfire navigation fundamentally different from conventional UAV navigation or indoor embodied-AI tasks, and demand systems that can \emph{perceive}, \emph{reason}, and \emph{act} under uncertainty while incorporating semantic understanding of wildfire phenomena, even relative to recent risk-aware navigation frameworks in 3D environments~\cite{khass2025active}.

To address these challenges, this work introduces an integrated simulation-to-autonomy pipeline that combines reinforcement learning (RL), vision--language semantic reasoning, and a physically grounded wildfire \textit{digital twin}. The digital twin encodes real-world terrain, vegetation fuels, wind fields, and fire propagation dynamics, enabling safe, large-scale training of UAV agents without endangering personnel or equipment. Building on recent digital-twin wildfire modeling research, our environment is constructed from USGS Digital Elevation Models (DEMs), LANDFIRE fuels, and semi-physical fire spread equations, allowing agents to interact with realistic slope-driven runs, plume occlusion, and wind-induced fire behavior \cite{coen2013cawfe,rollins2009landfire}.

Within this environment, we train a continuous-control UAV policy using proximal policy optimization (PPO), augmented with a reward component derived from a vision--language model (VLM) \cite{schulman2017ppo,radford2021learning}. The VLM evaluates UAV imagery using wildfire-specific textual prompts (e.g., ``active flame front,'' ``dense smoke column''), producing semantic alignment scores that reward visually meaningful observations. This fusion of physics-based RL incentives and high-level semantic guidance encourages the agent to maintain safe, information-rich viewpoints of the fire front, even in visually degraded conditions.

\noindent Our contributions are as follows:
\begin{enumerate}
    \item We construct a high-fidelity wildfire digital twin for UAV navigation, integrating terrain, fuel models, wind variability, and dynamic fire propagation.
    \item We develop a VLM-guided reinforcement learning framework that enhances UAV perception through semantic wildfire understanding.
    \item We introduce a wildfire-aware reward function that combines firefront proximity, plume visibility, hazard avoidance, and VLM-based semantic alignment.
    \item We demonstrate improved firefront localization accuracy, smoother trajectories, and stronger generalization across unseen wind regimes and fuel types.
\end{enumerate}

Overall, this work represents a step toward scalable, intelligent, autonomous wildfire monitoring systems capable of assisting incident response during real-world operations.

\section{Related Work}

Research on autonomous wildfire monitoring spans three primary areas: (1) UAV coordination for large-scale wildfire coverage, (2) supervised wildfire detection using RGB/IR data, and (3) simulation-based wildfire modeling and digital-twin systems. Our work bridges these areas by combining UAV autonomy, wildfire perception, and physics-grounded simulation in a single framework.

\subsection{UAV Coordination and Decentralized Wildfire Monitoring}

Early UAV wildfire monitoring systems emphasized decentralized coordination and perimeter coverage. Afghah et al.~\cite{afghah2019acc} proposed a multi–UAV architecture using distributed consensus to track dynamic firefronts while maintaining connectivity constraints, but under simplified dynamics and sensing assumptions. More recent work such as PyroTrack~\cite{khoshdel2024pyro} formulates belief-based RL for 2D UAV navigation in a POMDP grid environment, improving fire-line tracking over idealized fire maps without realistic fuels, wind, or coupled fire dynamics. Similarly, \emph{Distributed wildfire surveillance with autonomous aircraft using deep reinforcement learning}~\cite{shrestha2021distributed} applies DRL to coordinate multiple fixed-wing UAVs over simplified fire-intensity grids, again without a physically grounded fire–atmosphere model or high-fidelity 3D sensing.

Other decentralized frameworks explore partitioned search, multi-tier sensing, and consensus-based fireline reconstruction~\cite{pham2013distributed,casbeer2006cooperative,ollero2006uav}, but typically assume simplified sensing, kinematics, and fire dynamics. In contrast, our work embeds closed-loop UAV decision-making within a wildfire digital twin that models terrain, fuels, wind variability, and dynamic fire spread, and couples this environment with a VLM-guided RL policy operating directly on RGB observations.

\subsection{RGB/IR Datasets for Supervised Wildfire Detection}

Supervised wildfire perception has been driven by a limited set of UAV-based RGB/IR datasets, collected primarily over prescribed or pile burns rather than large, uncontrolled wildfires \cite{flame2paper,wilson2022wildfirenet,kou2020flame,gaetano2015wildfire}. These datasets provide synchronized RGB/thermal imagery with flame and smoke annotations, enabling CNN-based hotspot and wildfire detection. However, they are limited in scale, geographic and fuel diversity, and typically omit agent trajectories, action labels, and coupled terrain–wind–fire models, precluding embodied decision-making, long-horizon planning, or reasoning about where to move next.

Our VLM-guided RL framework addresses these limitations by training agents inside a wildfire-aware digital twin and using semantic alignment from a VLM to inject high-level wildfire concepts (e.g., flame fronts and smoke plumes) directly into the reward signal.

\subsection{Wildfire Digital Twins}
\begin{figure}[t]
\centering
\includegraphics[width=1\linewidth]{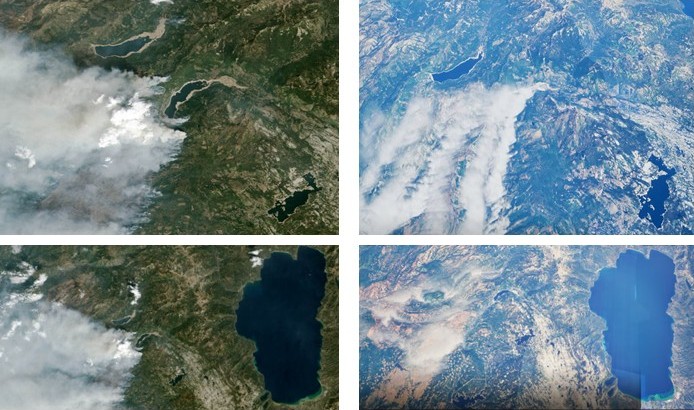}
\caption{King Fire side-by-side comparison. Natural-color Landsat 8 imagery of the fire is displayed in the left column as a wider satellite view at the bottom and a close-up at the top \cite{nasa_king_fire_image}. The equivalent digital twin-based aerial reconstruction produced using CAWFE simulation results is displayed in the right column as (top) close-up and (bottom) satellite view.}
\label{fig:KingFire}
\end{figure}


The concept of a wildfire \emph{digital twin}—a virtual environment grounded in GIS terrain, fuel models, and physically based fire–atmosphere interactions—has gained traction for situational awareness and decision support \cite{mahmud2022wildfireDT, fernando2021wildfireDT}. Recent systems couple numerical fire models such as CAWFE or WRF-SFIRE with GIS pipelines to produce high-fidelity forecasts and visual analytics \cite{coen2013modelling, kondratenko2020wrfsfire}. In this spirit, we construct physics-driven digital twins of real wildfire incidents using observed weather, LANDFIRE fuels, and NASA/USGS DEMs, and replay or re-simulate these events for UAV training (Fig.~\ref{fig:KingFire}).

For RL-based UAV control, satellite products alone are insufficient, as they do not provide height-resolved wind fields, plume-driven updrafts, temperature and heat-flux distributions, or spatially varying fuel consumption. Physics-based fire models supply these quantities at minute-scale timesteps on structured grids, which can be streamed into Unreal Engine~5 to drive realistic fire visualization, smoke occlusion, and dynamic lighting.

Most existing wildfire digital twin systems focus on visualization, fire progression analysis, or operator-facing decision support \cite{clark2022operationalDT}. They rarely embed closed-loop UAV policies interacting directly with the fire simulation, and, to our knowledge, none integrate reinforcement learning with VLM-guided semantic reward shaping within a wildfire-specific twin. Our framework addresses this gap by treating the digital twin as a training ground for embodied agents that must reason jointly about physical fire dynamics and their visual, semantic signatures.

\subsection{RL and Semantic Guidance in Embodied AI}

Reinforcement learning has shown strong performance in embodied navigation tasks such as exploration and obstacle avoidance, but wildfire environments introduce additional challenges, including rapidly changing visibility, wind disturbances, and fire-induced updrafts. Existing RL-based UAV navigation studies typically operate in static or simplified environments and lack semantic grounding \cite{bojarski2016end, sutton2018rl, lillicrap2015ddpg}. While embodied-AI benchmarks such as Habitat and Habitat~2.0~\cite{savva2019habitat, szot2021habitat2} emphasize high-fidelity simulation, they focus on indoor scenes with largely static obstacles rather than large-scale, partially observable landscapes with evolving fire dynamics.

Vision–language models (VLMs) such as CLIP, BLIP, and LLaVA provide strong zero-shot semantic reasoning in both static and embodied settings \cite{radford2021learning, li2022blip, liu2024llava}. Recent work has begun exploring VLM guidance for navigation and exploration \cite{shen2024nvila, habibpour2025think}, but applications to wildfire imagery remain limited. Our work bridges this gap by integrating wildfire-relevant VLM semantic alignment directly into the PPO reward function, encouraging UAV policies to favor observations that correspond to meaningful wildfire structures while operating in a physics-grounded digital twin.


\section{System Overview}
\label{sec:system_overview}

Our goal is to learn a wildfire-aware UAV monitoring policy that operates in a high-fidelity digital twin, combining physics-based simulation, reinforcement learning, and vision–language guidance. The system couples the wildfire digital twin with a Cosys-AirSim multirotor agent, a PPO-based policy, and a CLIP-style VLM that shapes the reward signal and provides directional guidance toward active fire fronts.


\subsection{Digital Twin Environment}
\label{subsec:dt_air}

The simulation stack is built around the digital twin environment, implemented in Unreal Engine~5.3, which provides:

\begin{itemize}
    \item high-resolution 3D terrain generated from USGS DEMs, with LANDFIRE fuel maps and CAWFE-derived fireline products.
    \item spatially varying fuel properties and burn progression.
    \item time-varying wildfire perimeters and intensity fields.
    \item configurable environmental conditions (e.g., wind, smoke, visibility).
\end{itemize}

We additionally integrate the AirSim framework to provide physically consistent multirotor dynamics, sensor simulation, and a standard RL interface. The AirSim UAV agent is spawned inside each scene and controlled through the AirSim API. At each time step, AirSim updates UAV pose and velocities, applies wind and gravity, and renders the top-down and angled cameras directly from the Unreal scene. 



\subsection{POMDP Formulation and UAV Control}
\label{subsec:obs}

We formulate the wildfire monitoring problem as a partially observable Markov decision process (POMDP) \cite{kaelbling1998planning}, where the UAV primarily observes camera imagery along with a small set of kinematic state variables and must infer latent firefront dynamics. A PPO agent \cite{schulman2017ppo,schulman2016gae} operates over a small discrete set of motion primitives \cite{anderson2018vision}, issuing one of seven direction commands (Forward, Right, Downward, Backward, Left, Upward, Hover) at each timestep. The environment maps these discrete actions to low-level multirotor control inputs in the simulator. We implement and train all agents using Stable-Baselines3 \cite{raffin2021stable}. We define:
\begin{equation}
    \mathcal{M} = (S, A, P, R, O, \gamma),
\end{equation}
where $S$ is the true simulator state - including the UAV position \& orientation, $A$ is the continuous–discrete hybrid action space, $P$ is the transition dynamics implemented by AirSim, $R$ is the reward function, $O$ is the observation space, and $\gamma$ is the discount factor.

At each timestep t, the UAV agent receives its position at the current timestep $\text{pos}_t$, its position at the previous timestep $\text{pos}_{t-1}$, its current velocity $v_t$, and its collision state.  The current \& previous position are both provided as XYZ coordinates using AirSim's NED (North, East, Down) coordinate system, mimicking a UAV's onboard GPS or IMU.  Similarly, the velocity is provided in the XYZ directions in the same format as the position.  The collision state is provided simply as a boolean True/False variable, indicating if the UAV has collided with another object.

Additionally, at timestep $t$, the observable state $o_t \in O$ includes:
\begin{itemize}
    \item a top-down RGB image $I^{\text{top}}_t$,
    \item a $45^\circ$-angled RGB image $I^{\text{ang}}_t$,
\end{itemize}

The UAV is controlled using a 7-dimensional discrete action space,
 \[
\begin{aligned}
A_{\text{move}} = \{ &\text{Forward = ($+v_x$, 0, 0), Right = (0, $+v_y$, 0),} \\                   &\text{Downward = (0, 0, $+v_z$), Backward = ($-v_x$, 0, 0),} \\              &\text{Left = (0, $-v_y$, 0), Upward = (0, 0, $-v_z$),} \\
                     &\text{Hover = (0, 0, 0)}\}.
\end{aligned}
\]

\noindent where each action corresponds to a canonical motion primitive in the local UAV frame. At each timestep $t$, the policy outputs a discrete action index $a_t \in \{0,\dots,6\}$. The environment maps this index to a fixed-magnitude velocity command along the corresponding axis (or zero velocity for \emph{hover}) and passes it to the low-level multirotor controller in the simulator. This design keeps the action space small and interpretable while still allowing the agent to express rich 3D tracking behaviors around the wildfire front.


\paragraph{Base Reward Design.}
\label{subsec:base_reward}

Before adding the VLM term, the PPO agent is trained with a base reward that balances exploration, energy efficiency, safe altitude, and collision avoidance. At time $t$ the base reward is
\begin{equation}
    R_{\text{base},t}
    = w_m R_m(t) + w_e R_e(t) + w_z R_z(t) + R_c(t),
\end{equation}
where $w_m = 0.1$, $w_e = -0.001$, $w_z = -0.0005$.

The movement term, $R_m$  encourages both local motion and exploration away from the deployment point.
This term prevents the agent from hovering in place and promotes continual coverage of new areas. The energy term $R_e$
 penalizes aggressive maneuvers using a simple multirotor power model~\cite{liu2017power}.
\begin{equation}
    R_e(t) = E_t
    = \sum \bigl(mg|\dot{z}| + 0.5 \rho v^3 C_d A\bigr)\,\Delta t \,\times \beta,
\end{equation}

\begin{equation}
    \beta =
    \begin{cases}
        1.0, & \text{hover},\\
        1.2, & \text{normal flight},\\
        1.6, & \text{aggressive flight}.
    \end{cases}
\end{equation}
where $m$ is UAV mass, $g$ gravity, $\rho$ air density, $v$ airspeed, $C_d$ drag coefficient, $A$ reference area, and $\beta$ is a mode-dependent multiplier. To encourage rapid detection, the energy penalty is applied only until wildfire is detected in the top-down RGB view, after which $R_e$ is set to zero.

To discourage unrealistic high-altitude “zoom-out” behavior, we define an altitude penalty $R_z$ that penalizes altitude above 400\,m AGL:
\begin{equation}
    R_z(t) =
    \begin{cases}
        z_t, & z_t > 400 \text{ m AGL},\\
        0,   & \text{otherwise},
    \end{cases}
\end{equation}
This keeps the UAV at altitudes that preserve useful image resolution and limit unnecessary energy use.
The final term prioritizes safe flight; in practice, a collision both applies a large negative reward $R_c$ and terminates the episode.

\subsection{VLM-Guided Reward Shaping}
\label{subsec:vlm_shaping}

A key component of the system is a pretrained VLM that shapes the reward signal based on wildfire semantics. We use a CLIP-style encoder (ViT-B/32) \cite{openaiclip} operating on two RGB views: a $45^\circ$ angled horizon image and a nadir (top-down) image. The VLM is queried every few timesteps (frame\_skip $=4$ in our implementation), and embeddings are reused between queries to reduce computational cost.

\paragraph{Angled-view potential shaping.}
The angled RGB image $I^{\text{ang}}_t$ provides long-range context (smoke plumes, distant flames) that may not yet be visible in the top-down view. At each VLM update, we compute the cosine similarity between the two embeddings corresponding to the angled RGB image $I^{\text{ang}}_t$ and a fixed wildfire text prompt, \textit{``an aerial image of an active wildfire with smoke plumes''}. The resulting scalar score, denoted $s_t$, indicates how closely the current view matches the target wildfire concept.

To incorporate semantic guidance into the PPO loop without altering the optimal policy, the VLM similarity score is treated as a state potential function rather than a direct reward.  This follows potential-based reward shaping methods, where the reward from the auxiliary signal depends only on the change in a potential $\phi(s)$ across consecutive states.  In this setting, the CLIP similarity score $s_t \in [-1,1]$ is interpreted as a potential describing how “wildfire-relevant" the current RGB image observation is.  The PPO agent then receives a reward only when the agent increases wildfire relevance over time, ensuring that the shaping term accelerates learning without modifying the underlying optimal policy.

This score is mapped from $[-1,1]$ to a centered potential:
\begin{equation}
    \phi_t = \text{scale} \left( \frac{s_t + 1}{2} - 0.5 \right),
\end{equation}
where the scalar \texttt{scale} controls the magnitude of the shaping signal. The corresponding potential-based shaping reward is
\begin{equation}
    R_{\text{VLM}}(t) = \gamma \, \phi_t - \phi_{t-1},
\end{equation}
with $\gamma$ equal to the PPO discount factor. This term rewards trajectories that monotonically increase semantic alignment with wildfire, helping the agent lock onto the fire front even when it starts far from the burn area.

\begin{figure}[t]
\centering
\begin{minipage}{0.48\linewidth}
    \centering
    \includegraphics[width=\linewidth]{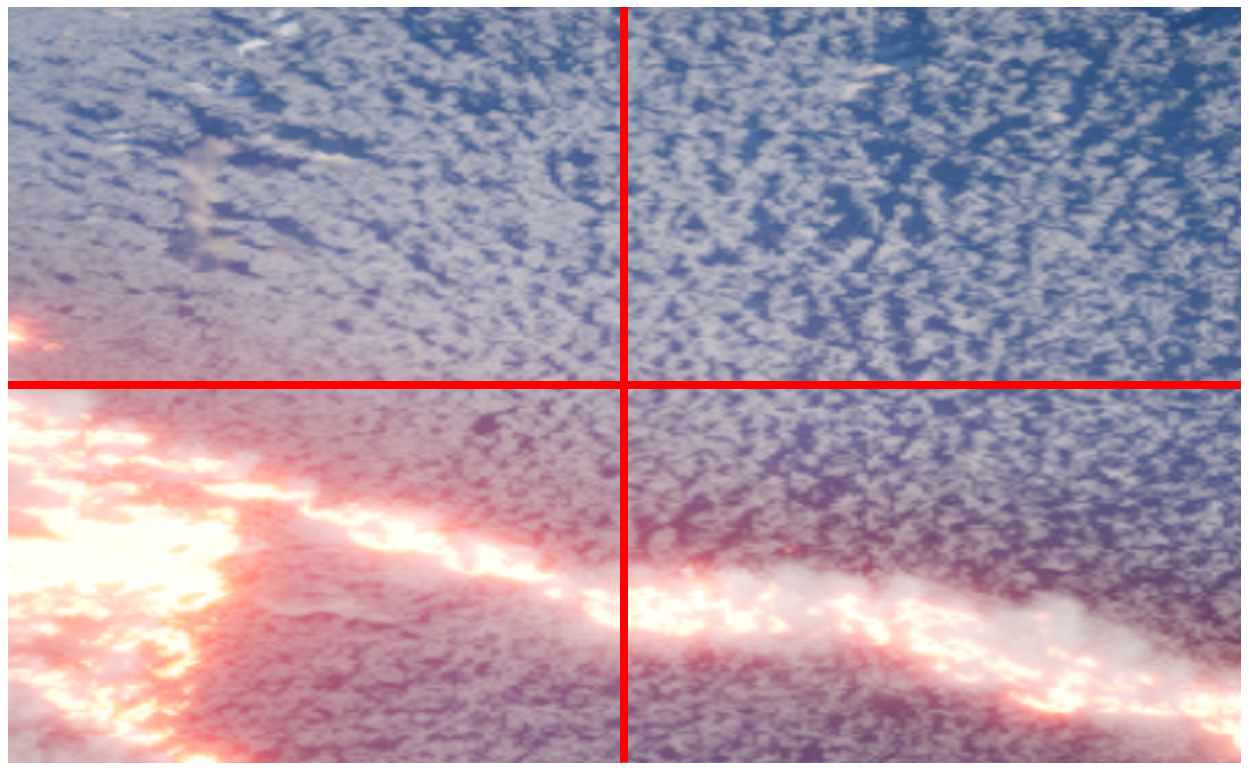}
\end{minipage}
\hfill
\begin{minipage}{0.48\linewidth}
    \centering
    \includegraphics[width=\linewidth]{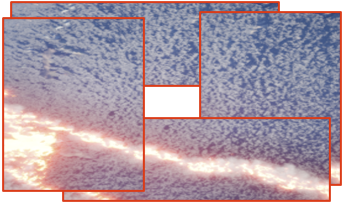}
\end{minipage}

\begin{minipage}{0.48\linewidth}
    \centering
    \includegraphics[width=\linewidth]{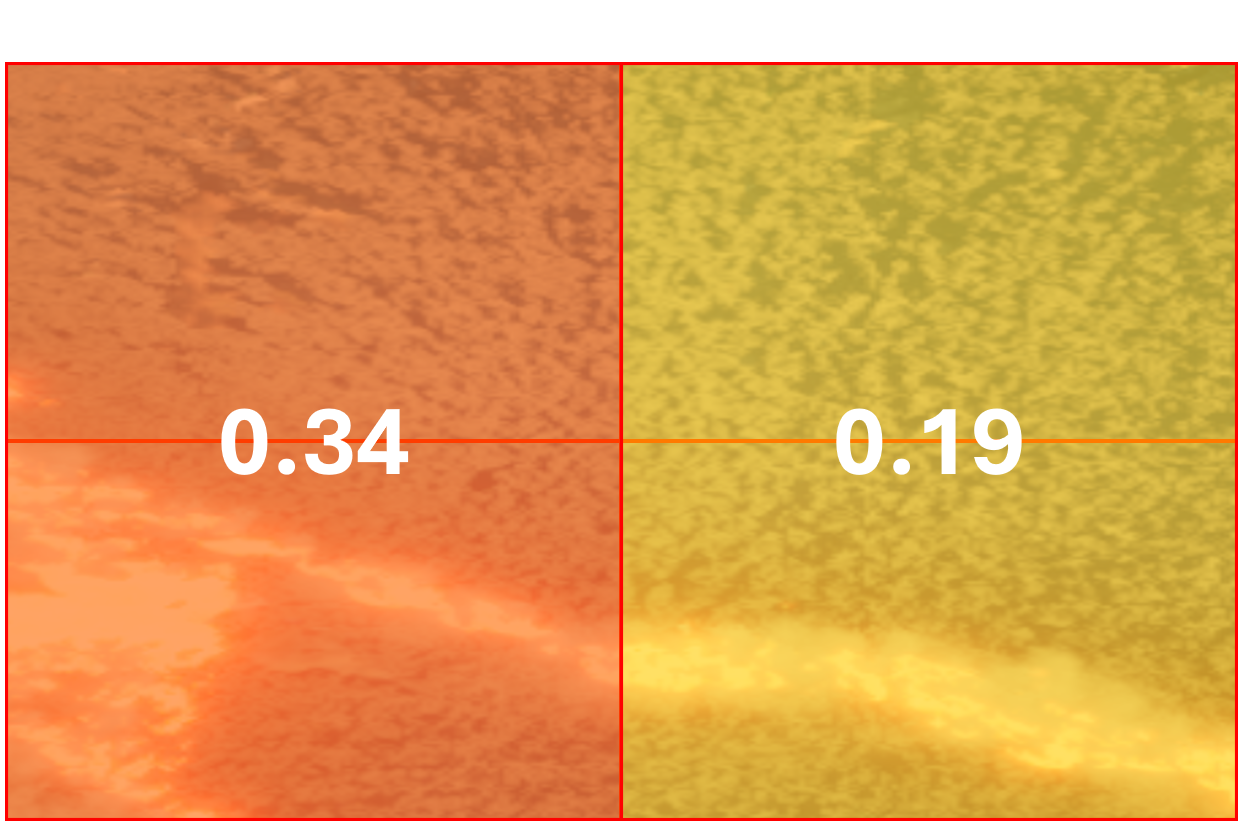}
\end{minipage}
\hfill
\begin{minipage}{0.48\linewidth}
    \centering
    \includegraphics[width=\linewidth]{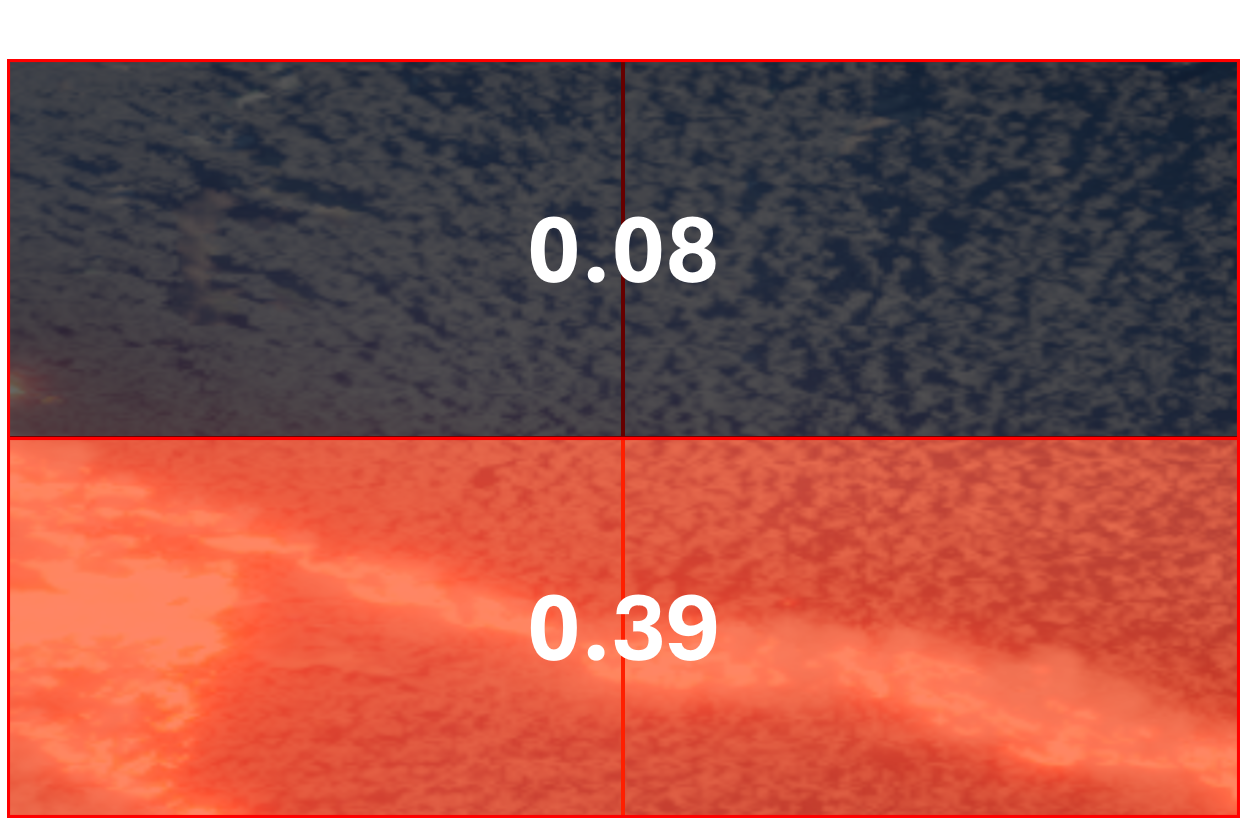}
\end{minipage}
\caption{Top left: top-down UAV camera view with red crosshairs indicating the four directional quadrants (forward, backward, left, right).  Top right: corresponding cropped patches extracted from each quadrant, which are passed independently to the VLM to estimate wildfire likelihood in each direction.  Bottom left: VLM directional guidance for left \& right patches.  Bottom right: VLM directional guidance for forward \& backward patches.}
\label{fig:topdown_quadrants}
\end{figure}

\paragraph{Top-down directional guidance.}
To provide local navigation hints, the top-down RGB image $I^{\text{top}}_t$ is partitioned into four directional patches corresponding to the UAV’s forward, backward, left, and right quadrants, as illustrated in Fig.~\ref{fig:topdown_quadrants}. For each quadrant, we compute the cosine similarity with the wildfire text prompt and map these similarities to non-negative scores, which are then normalized to form a directional probability distribution $P_d$ over $d \in \{f,b,l,r\}$. This distribution can be interpreted as a semantic heatmap over directions. Let
\begin{equation}
    d^* = \arg\max_{d \in \{f,b,l,r\}} P_d
\end{equation}
denote the most likely direction of wildfire activity according to the VLM. If the agent’s chosen action direction $a_t$ matches $d^*$, it receives a small directional bonus $\beta \, P_{d^*}$.
Where the coefficient $\beta$ is set to $0.05$ so that this term remains modest relative to the base reward. This encourages the UAV to move toward quadrants whose appearance is semantically closest to active wildfire.

The final scalar reward used for PPO optimization combines physical and semantic terms:
\begin{equation}
    R_{\text{total}}(t) = R_{\text{base},t} + R_{\text{VLM}}(t) + R_{\text{VLM-d}}(t).
\end{equation}
In practice, the scaling factor in $\phi_t$ and the coefficient $\beta$ are chosen so that both VLM-based components have magnitudes comparable to the base reward $R_{\text{base},t}$. This normalization prevents any single term from dominating, while allowing wildfire-relevant visual cues from both the angled and top-down views to meaningfully shape the learned policy.

\begin{algorithm}[t]
\caption{VLM-Guided UAV Wildfire Monitoring}
\label{alg:vlm_ppo}
\footnotesize
\begin{algorithmic}[1]
\Statex

\State Compute text embedding $z_T \gets f_T(T_{\text{fire}})$

\For{ training iteration}
    \While{episode not done and $t < T_{\text{max}}$}
        \State $o_t \gets \{I^{\text{top}}_t, I^{\text{ang}}_t\}$
        \State $a_t \sim \pi_\theta(\cdot \mid o_t)$
        \State $\mathbf{v}_t \gets \mathrm{MotionPrimitive}(a_t)$
        \State $(I^{\text{top}}_{t+1}, I^{\text{ang}}_{t+1}, s^{\text{UAV}}_{t+1}, c_{t+1}, s^{\text{fire}}_{t+1}) \gets \mathcal{P}(s_t, \mathbf{v}_t)$

        \State \textbf{ Base reward computation $R_{\text{base},t}$}
        \State $R_{\text{base},t} \gets w_m R_m(t) + w_e R_e(t) + w_z R_z(t) + R_c(t)$

        \State \textbf{ VLM-guided rewards}
        \If{$t \bmod \texttt{frame\_skip} = 0$}
            \State $z_I \gets f_I(I^{\text{ang}}_{t+1})$
            
            \State $s_t = \mathrm{cosSim}(z_I, z_T)$

            \State  $\phi_t \gets \text{scale} \left(\frac{s_t + 1}{2} - 0.5\right)$
            \State $R_{\text{VLM}}(t) \gets \gamma \phi_t - \phi_{\text{t-1}}$
            
            \State Partition $I^{\text{top}}_{t+1}$ into quadrants $\{i_f, i_b, i_l, i_r\}$
            \For{ $d \in \{f,b,l,r\}$}
                \State $e^{(d)} \gets f_I(i_d)$
                \State $s^{(d)} = \mathrm{cosSim}(e^{(d)}, z_T)$
                
                \State $\tilde{s}^{(d)} \gets \frac{s^{(d)} + 1}{2}$
            \EndFor
            \State $P_d = \frac{\tilde{s}^{(d)}}{\sum_{k} \tilde{s}^{(k)}}, \qquad \forall d.$
        \Else
            \State $R_{\text{VLM}}(t) \gets R_{\text{VLM}}(t-1)$
            \State $P_d(t) \gets P_d(t-1), \quad \forall d \in \{f,b,l,r\}$

        \EndIf

        \State $d^* \gets \arg\max_{d} P_d$
        \State $\hat{d}_t \gets g(a_t)$ \Comment{$g(\cdot)$ maps action index $a_t$ to a direction label}
        \State $R_{\text{VLM-d}}(t) \gets \mathbb{1}[\hat{d}_t = d^*] \, \beta \, P_{d^*}$


        \State $R_{\text{total}}(t) \gets R_{\text{base},t} + R_{\text{VLM}}(t) + R_{\text{VLM-d}}(t)$
        \State Store $(o_t, a_t, R_{\text{total}}(t), o_{t+1}, \text{done})$ in buffer $\mathcal{D}$
        \State $t \gets t + 1$
    \EndWhile

    \State \textbf{PPO update (GAE + clipped objective)}
    \State $\{\hat{A}_t, \hat{R}_t\}_{t=0}^{T-1} \gets \mathrm{GAE}(\mathcal{D})$
    \State $(\theta, \psi) \gets \mathrm{UpdatePPO}(\mathcal{D})$
    \State $\mathcal{D} \gets \emptyset$

\EndFor

\end{algorithmic}
\end{algorithm}

\subsection{Training Pipeline}
\label{subsec:ppo_pipeline}

We train a PPO agent using the Stable-Baselines3 implementation \cite{sb3ppo} with a multi-input policy (MultiInputPolicy). The policy network ingests a concatenation of CNN features from the top-down and angled RGB images together with scalar state features (e.g., altitude, velocity, and other kinematic variables). All PPO hyperparameters follow Table~\ref{tab:hyperparameters}.

Training is performed for a total of 200{,}000 timesteps with 4{,}000 timesteps per episode. An evaluation callback is triggered every 20{,}000 timesteps, yielding 50 training episodes and 10 evaluation runs in total; the best-performing checkpoint (by evaluation return) is saved at each evaluation.

\noindent At a high level, each PPO iteration proceeds as follows:
\begin{enumerate}
    \item \textbf{Rollout:} Collect $T$ steps of experience in the digital twin environment using the current policy.
    \item \textbf{Reward computation:} For each step, compute $R_{\text{base},t}$, then use the VLM wrapper to obtain the potential-based term $R_{\text{VLM}}(t)$ and the directional term $R_{\text{VLM-d}}(t)$, forming the total reward $R_{\text{total}}(t)$
   
    \item \textbf{Advantage estimation:} Compute advantages $\hat{A}_t$ using generalized advantage estimation (GAE) with discount factor $\gamma$ and parameter $\lambda$:
    \begin{equation}
        \hat{A}_t = \sum_{l=0}^{T-t} (\gamma \lambda)^l \, \delta_{t+l},
    \end{equation}
    where the temporal-difference residual is
    \begin{equation}
        \delta_t = R_{\text{total}}(t) + \gamma V(s_{t+1}) - V(s_t).
    \end{equation}
    \item \textbf{Policy update:} Optimize the clipped PPO objective with value and entropy regularization:
    \begin{equation}
        \mathcal{L}_{\text{total}} =
        \mathcal{L}_{\text{clip}} +
        c_v \mathcal{L}_{\text{value}} +
        c_h \mathcal{L}_{\text{entropy}}.
    \end{equation}
\end{enumerate}

Overall, this pipeline couples the digital twin, physically based UAV simulation, PPO-based RL, and VLM-guided reward shaping into a single end-to-end loop, enabling the agent to learn monitoring behaviors that are both physically plausible and semantically aligned with active wildfire fronts.

\section{Experimental Results and Evaluation}
\label{sec:results}




Because, to our knowledge, there is no prior VLM–RL framework for UAV-based wildfire monitoring, we evaluate our method through a controlled ablation study rather than comparison to external baselines. While simple heuristic search baselines (e.g., spiral or lawnmower patterns) could be added, integrating them fairly into the same physics-grounded digital twin stack is left to future work; here we focus on disentangling the impact of VLM guidance and dual-view sensing within a controlled PPO-based ablation. We ablate models that are trained with the same Stable-Baselines3 PPO pipeline and hyperparameters, and are evaluated under identical initial conditions and trajectories. The variants differ only in architectural components such as the presence of VLM guidance, use of top-down vs.\ angled cameras, and the directional segmentation module, allowing us to isolate the contribution of each subsystem.

For quantitative evaluation, we report several key performance indicators over a 4000-timestep episode: (i) \textbf{Total Reward} ($R_{\text{total}}$), the cumulative return; (ii) \textbf{\% Time in FOV}, the fraction of the episode during which wildfire is visible in the top-down camera; (iii) \textbf{Time to Detection (TTD, sec)} and \textbf{Timesteps to Detection}, defined as the first time the wildfire pixel count exceeds a fixed threshold; (iv) \textbf{Total Distance (m)} traveled in the $x$–$y$ plane, which reflects exploration efficiency; and (v) \textbf{Runtime (sec)}, which measures end-to-end evaluation time. Together, these metrics capture detection speed, monitoring quality, and path efficiency, as well as robustness across different environmental conditions.

Each model is evaluated on five wildfire monitoring tasks of increasing difficulty: (1) fire already in the initial top-down field of view (no search required, as sanity check); (2) fire near the UAV but initially outside the FOV; (3) fire approximately 1\,km away with only a distant smoke cue; (4) near-fire with high wind ($\approx 5$\,m/s) to test robustness to drift; and (5) near-fire under low-light conditions to emulate nighttime operation. Tasks 2–5 require varying degrees of exploration, semantic reasoning, and disturbance rejection, while Task~3 in particular stresses long-range search under sparse visual cues.

\begin{figure*}[t]
    \centering
    \begin{subfigure}[b]{0.23\linewidth}
        \centering
        \includegraphics[width=\linewidth]{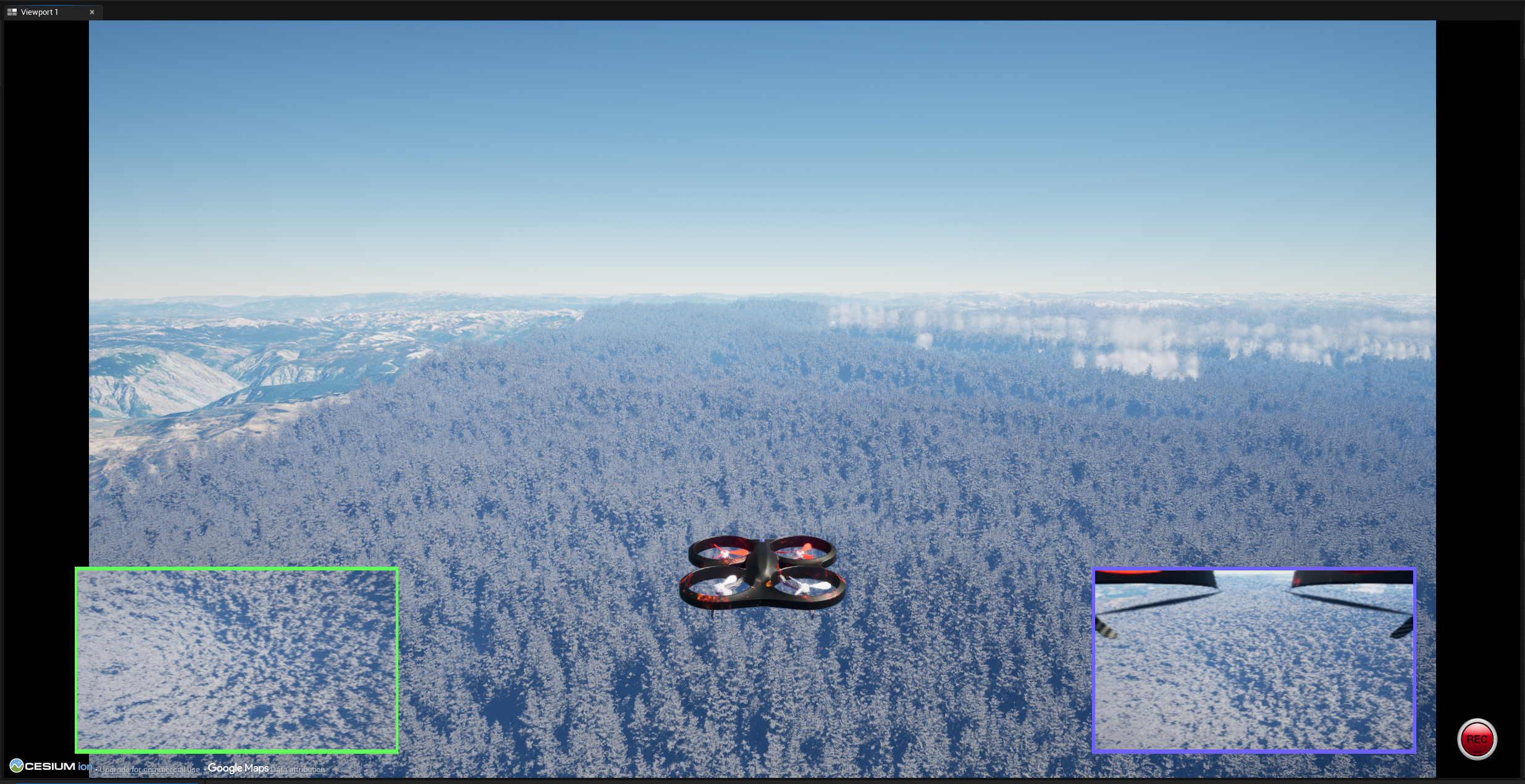}
        \caption{Initial search}
        \label{fig:progression_search}
    \end{subfigure}
    \hfill
    \begin{subfigure}[b]{0.23\linewidth}
        \centering
        \includegraphics[width=\linewidth]{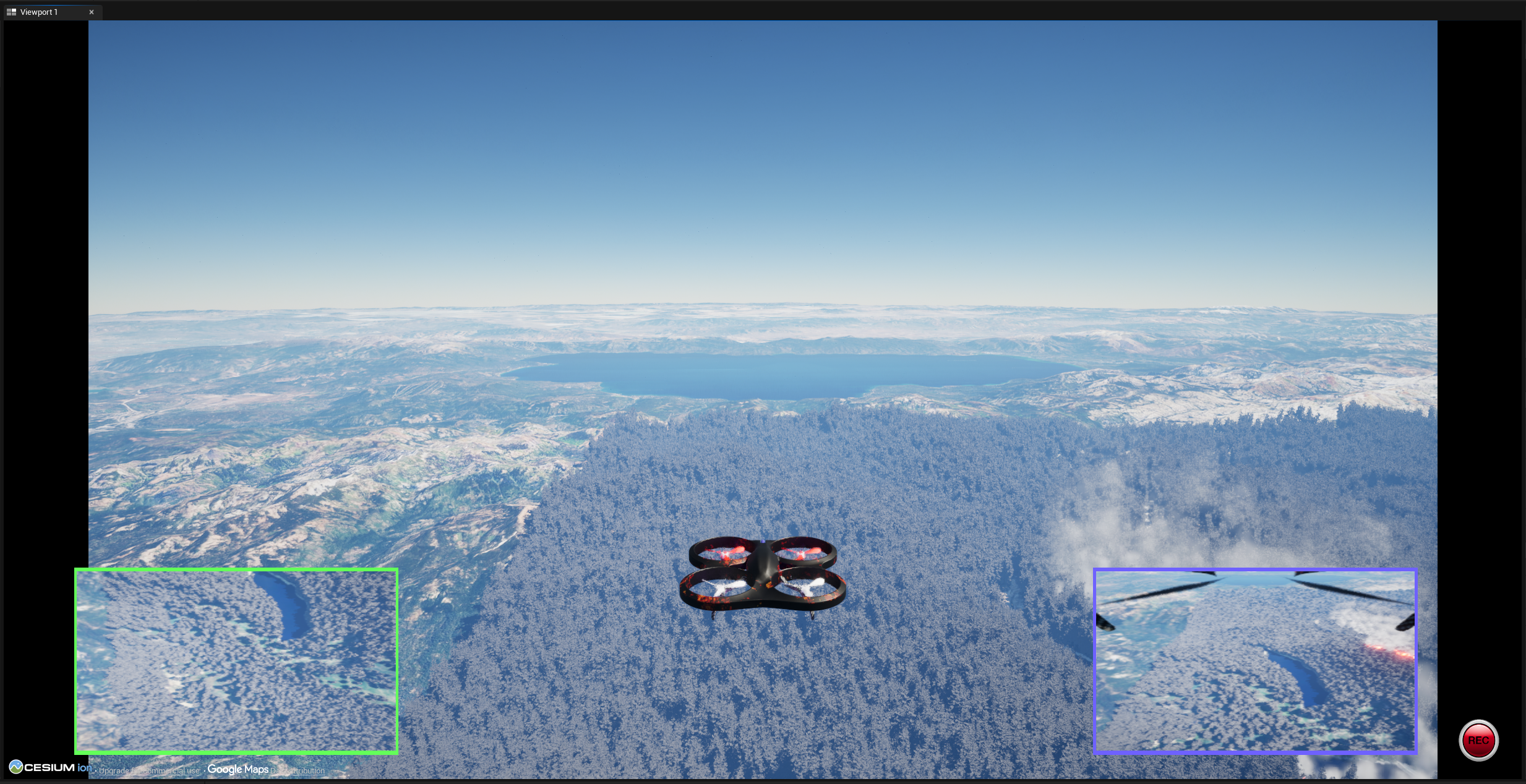}
        \caption{First smoke cue}
        \label{fig:progression_detection}
    \end{subfigure}
    \hfill
    \begin{subfigure}[b]{0.23\linewidth}
        \centering
        \includegraphics[width=\linewidth]{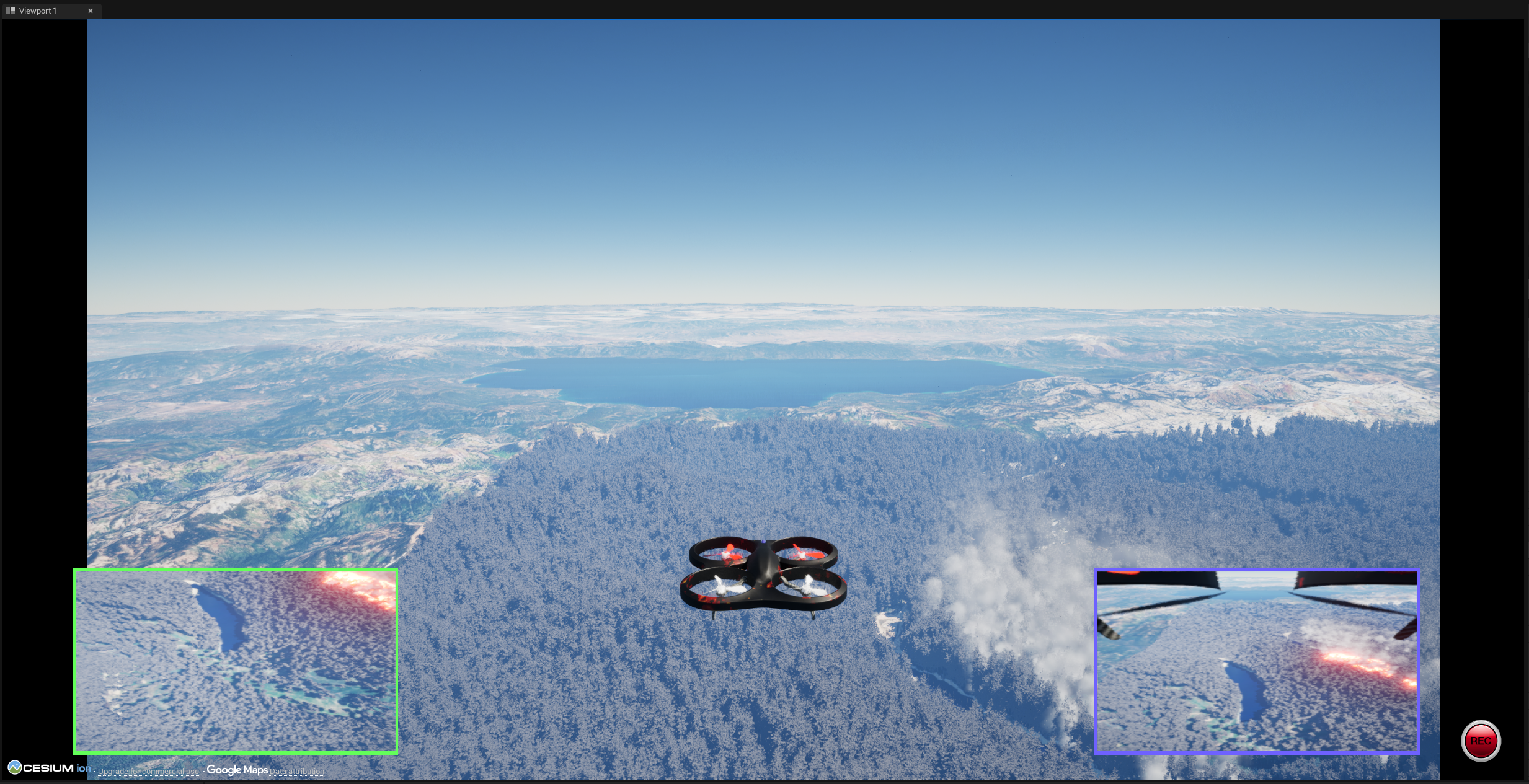}
        \caption{Locked on firefront}
        \label{fig:progression_lock}
    \end{subfigure}
    \hfill
    \begin{subfigure}[b]{0.23\linewidth}
        \centering
        \includegraphics[width=\linewidth]{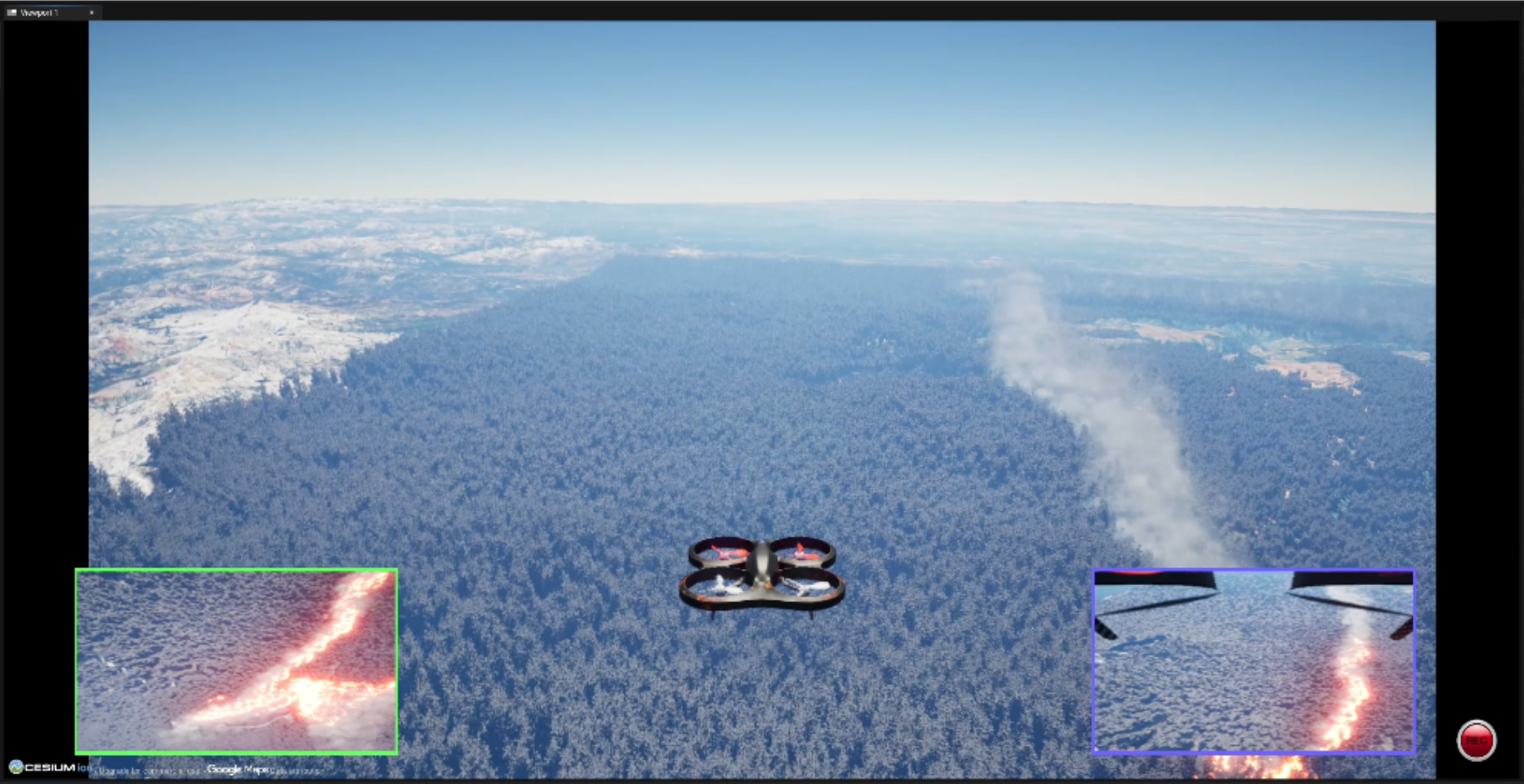}
        \caption{Sustained tracking}
        \label{fig:progression_tracking}
    \end{subfigure}

    \caption{
    Qualitative episode showing the progression of the VLM-guided UAV policy from exploration to sustained tracking. 
    Each panel shows a composite of the top-down (top) and angled (bottom) RGB views at a representative timestep:
    (a) initial search with no wildfire in the FOV, 
    (b) first distant smoke cue used by the VLM to steer the agent, 
    (c) locked-on view of the active firefront, and 
    (d) sustained tracking while following the fire perimeter.
    }
    \label{fig:episode_progression}
\end{figure*}

\subsection{Ablation Study: Near-Fire Search (Task 2)}

Table~\ref{tab:ablation2_main} reports results for Task~2, where the UAV starts $\approx 100$\,m from the wildfire with no fire in its initial top-down view. This scenario requires short-range search followed by tracking.


\begin{table}[t]
\begin{center}
\centering
\renewcommand{\arraystretch}{1.5}
\setlength{\tabcolsep}{2.0pt}
\resizebox{\columnwidth}{!}{
\begin{tabular}{|c|c|c|c c|c|c|}
\hline
\textbf{Model} &
\textbf{$R_{\text{total}}$} &
\textbf{\%FOV} &
\textbf{TTD (s)} &
\textbf{(steps)} &
\textbf{Dist (m)} &
\textbf{Runtime (s)} \\
\hline
\textbf{Base PPO} &
$4371.6 \pm 69.9$ &
$86.14 \pm 1.06$ &
$641.42 \pm 21.92$ &
$574.2 \pm 19.1$ &
$3903.3 \pm 85.6$ &
$\mathbf{4513.5} \pm 14.0$ \\
\hline
\textbf{VLM-only} &
$3587.6 \pm 57.5$ &
$89.74 \pm 0.64$ &
$473.08 \pm 15.66$ &
$415.0 \pm 13.2$ &
$2057.0 \pm 56.8$ &
$4588.3 \pm 7.7$ \\
\textbf{rew. shaping} & & & & & & \\
\hline
\textbf{VLM-int.,} &
$4724.4 \pm 48.9$ &
$91.78 \pm 0.54$ &
$319.44 \pm 12.26$ &
$271.0 \pm 9.7$ &
$5281.3 \pm 88.0$ &
$4701.7 \pm 4.0$ \\
\textbf{top-down only} & & & & & & \\
\hline
\textbf{VLM-int.,} &
$4007.5 \pm 79.0$ &
$89.12 \pm 0.80$ &
$359.09 \pm 13.18$ &
$278.8 \pm 9.9$ &
$5509.7 \pm 69.5$ &
$4710.0 \pm 2.1$ \\
\textbf{angled only} & & & & & & \\
\hline
\textbf{VLM-guided,} &
$5060.4 \pm 64.5$ &
$96.70 \pm 0.32$ &
$175.03 \pm 9.44$ &
$111.0 \pm 6.4$ &
$5801.2 \pm 65.6$ &
$4715.5 \pm 5.0$ \\
\textbf{unsegmented} & & & & & & \\
\hline
\textbf{VLM-guided} &
$\mathbf{5102.4} \pm 69.7$ &
$\mathbf{97.26} \pm 0.24$ &
$\mathbf{171.22} \pm 8.56$ &
$\mathbf{107.6} \pm 6.3$ &
$\mathbf{5816.6} \pm 75.7$ &
$4723.5 \pm 5.3$ \\
\textbf{final model} & & & & & & \\
\hline
\end{tabular}}
\caption{
Task 2 (wildfire near UAV $\approx 100$\,m away, initially outside FOV). Values are mean $\pm$ standard deviation over $N=5$ runs with different UAV spawning points inside the digital twin. The final VLM-guided model achieves the highest total reward, largest FOV coverage, and fastest detection while maintaining long, efficient trajectories.
}
\label{tab:ablation2_main}
\end{center}
\end{table}

When the fire is nearby but not directly visible, the final VLM-guided model yields the best overall performance: it achieves the highest cumulative reward, the highest time-in-FOV percentage of 97.1\%, and detects the fire in roughly one-third of the time required by the base PPO (180\,s vs.\ 669\,s). The VLM-guided model without directional segmentation performs similarly, confirming that the primary gains come from VLM-based semantic feedback and dual-view sensing, while the directional heatmap provides an additional efficiency boost. Baseline PPO and VLM-only reward shaping lag in both detection speed and FOV coverage, often relying on altitude increases instead of actively steering toward likely fire regions. A representative rollout for Task~2 is visualized in Fig.~\ref{fig:episode_progression}, where the VLM-guided policy transitions from an initial search phase with no fire in view, through first detection of a distant smoke plume, to a locked-in view of the active front and sustained perimeter tracking. This three-phase behavior is consistent with the quantitative gains in TTD and \%FOV, showing how early exploratory motion is rapidly converted into long-horizon, information-rich monitoring.

\subsection{Long-Range Search: Fire at 1 km (Task 3)}

Task~3 represents the most challenging scenario, with the UAV deployed approximately 1\,km from the fire and no wildfire in the initial top-down view. Only a faint smoke plume is visible on the angled camera, making rewards sparse and the search problem difficult. Table~\ref{tab:ablation3_main} summarizes the results.


\begin{table}[t]
\begin{center}
\centering
\renewcommand{\arraystretch}{1.5}
\setlength{\tabcolsep}{2.0pt}
\resizebox{\columnwidth}{!}{
\begin{tabular}{|c|c|c|c c|c|c|}
\hline
\textbf{Model} &
\textbf{$R_{\text{total}}$} &
\textbf{\%FOV} &
\textbf{TTD (s)} &
\textbf{(steps)} &
\textbf{Dist (m)} &
\textbf{Runtime (s)} \\
\hline
\textbf{Base PPO} &
$280.50 \pm 118.79$ &
$0.00 \pm 0.00$ &
\textbf{--} &
\textbf{--} &
$3902.76 \pm 228.41$ &
$\mathbf{4231.92} \pm 8.48$ \\
\hline
\textbf{VLM-only} &
$238.31 \pm 60.05$ &
$0.00 \pm 0.00$ &
\textbf{--} &
\textbf{--} & 
$1753.93 \pm 116.41$ &
$4362.94 \pm 11.08$ \\
\textbf{rew. shaping} & & & & & & \\
\hline
\textbf{VLM-int.,} &
$488.08 \pm 90.60$ &
$1.94 \pm 4.33$ &
$3444.18 \pm 383.56$ &
$3280.0 \pm 347.9$ &
$2653.92 \pm 518.59$ &
$4503.52 \pm 5.79$ \\
\textbf{top-down only} & & & & & & \\
\hline
\textbf{VLM-int.,} &
$548.78 \pm 58.23$ &
$2.70 \pm 6.05$ &
$3385.62 \pm 356.69$ &
$3192.0 \pm 317.0$ &
$3432.99 \pm 395.96$ &
$4506.28 \pm 5.93$ \\
\textbf{angled only} & & & & & & \\
\hline
\textbf{VLM-guided,} &
$2545.85 \pm 283.21$ &
$39.66 \pm 2.11$ &
$2868.66 \pm 137.17$ &
$2320.6 \pm 82.7$ &
$5086.76 \pm 340.67$ &
$4641.20 \pm 3.16$ \\
\textbf{unsegmented} & & & & & & \\
\hline
\textbf{VLM-guided} &
$\mathbf{3342.62} \pm 183.68$ &
$\mathbf{42.88} \pm 1.54$ &
$\mathbf{2687.21} \pm 129.37$ &
$\mathbf{2182.6} \pm 68.0$ &
$\mathbf{5483.70} \pm 301.09$ &
$4651.86 \pm 3.13$ \\
\textbf{final model} & & & & & & \\
\hline
\end{tabular}}
\caption{
Task 3 (wildfire $\approx 1$\,km away, outside initial FOV). Only the two dual-camera, VLM-guided models successfully detect and track the fire within the episode horizon. Each model is evaluated over $N=5$ runs with different UAV spawning points inside the digital twin; boldface indicates the best mean value in each column.
}
\label{tab:ablation3_main}
\end{center}
\end{table}

In this long-range setting, only the two dual-camera VLM-guided models successfully detect the wildfire within 4000 timesteps. The final model attains the highest total reward and maintains the fire in view for approximately 45\% of the episode, whereas all four baselines fail to detect the fire at all (\%FOV = 0). This sharp separation underscores the importance of combining (i) angled-view semantics for distant plume cues, (ii) top-down directional guidance, and (iii) potential-based VLM shaping for stabilizing exploration under extremely sparse rewards.

\subsection{Robustness to Wind and Low Light}

Tasks~4 and~5 reuse the “near-fire, out-of-FOV” setup from Task~2 but add high wind ($\approx 5$,m/s) and low-light conditions, respectively. In both cases, the VLM-guided final model achieves the highest total reward, the highest time-in-FOV (above 96\%), and the fastest detection, with only minor degradation relative to Task~2, while other models slow down more under wind and reduced visibility. This suggests that VLM-based semantic cues help compensate for noisy dynamics and lower image contrast. Full per-task tables and qualitative visualizations are provided in the Appendix.

Overall, the ablations show that (i) base PPO is adequate when the fire is already in view but fails once search is required; (ii) VLM reward alone is insufficient without a well-shaped physical base signal; and (iii) combining physics-aware rewards, dual-camera sensing, and VLM-guided potential and directional shaping is crucial for fast, robust wildfire localization and sustained front tracking across diverse conditions. All experiments were run on a workstation GPU (RTX 3060).

\subsection{Prompt  Ablation Study}
\label{subsec:prompt_ablation}

Because the VLM reward is derived solely from the cosine similarity between an image embedding and a text prompt, the choice of prompt has a direct impact on reward shaping. If the prompt is too short or ambiguous, the similarity score is noisy and weakly correlated with wildfire content. Conversely, overly specific or verbose prompts can hurt generalization, as small changes in viewpoint, lighting, or terrain cause a mismatch between the simulated imagery and the textual description.

To study this effect, we compared several candidate prompts for the CLIP-based reward, ranging from minimal descriptors to highly detailed phrases. Empirically, we found that the mid-level prompt
\textit{``an aerial image of an active wildfire with smoke plumes''}
produced the most stable and informative similarity signal across all evaluation scenarios. Shorter prompts tended to fire on non-wildfire bright regions (e.g., sunlit terrain), while very long prompts were brittle and less consistent across environments. Table~\ref{tab:prompt_ablation} summarizes the prompt variants and their observed behavior.

\begin{table}[t]
    \centering
    \scriptsize
    \renewcommand{\arraystretch}{1.3}
    \setlength{\tabcolsep}{2.0pt}
    \begin{tabular}{|c|p{0.22\linewidth}|p{0.12\linewidth}|p{0.47\linewidth}|}
        \hline
        \textbf{ID} & \textbf{Text Prompt} & \textbf{ $R_{\text{total}}$} & \textbf{Observed Behavior} \\
        \hline
        P1 &
        \textit{``wildfire''} & 4765.61 &
        Very high-level concept; cosine similarity often responds to generic bright or high-contrast regions, leading to noisy rewards and weak correlation with actual firefront location. \\
        \hline
        P2 &
        \textit{``an aerial image of a wildfire''} & 4975.23 &
        More grounded in aerial context and generally more stable than P1, but still less sensitive to smoke structure; tends to under-reward distant plumes and partially occluded fire. \\
        \hline
        P3 (ours) &
        \textit{``an aerial image of an active wildfire with smoke plumes''} & 5018.22 &
        Best trade-off between specificity and robustness. Consistently highlights flame edges and smoke columns at multiple ranges, providing a smooth, informative similarity signal for both angled and top-down views. \\
        \hline
        P4 &
        \textit{``a UAV image of a distant wildfire burning across a mountain ridge at night with thick smoke''} & 4920.84 &
        Overly specific; performs well only when the scene closely matches the full description. Similarity degrades when terrain, time of day, or viewpoint differ, reducing the usefulness of the VLM reward and harming generalization. \\
        \hline
    \end{tabular}
    \caption{Prompt ablation for the CLIP-based reward for Task 2. Extremely short prompts (P1) are too ambiguous, while overly detailed prompts (P4) are brittle and environment-specific. The adopted prompt (P3) provides the most reliable semantic signal for cosine-similarity-based reward shaping.}
    \label{tab:prompt_ablation}
\end{table}


\section{Conclusion}

This work introduced a unified framework that integrates VLM-guided semantic reasoning with PPO-based reinforcement learning inside a high-fidelity wildfire digital twin. By combining physics-informed fire simulation, GIS-derived terrain and fuel modeling, multimodal UAV sensing, and VLM-based semantic alignment, our system enables robust autonomous tracking of wildfire fronts under challenging visibility and wind conditions. Across a suite of evaluation tasks and ablated variants, the proposed approach improves firefront localization, increases time-in-FOV, and accelerates detection compared to purely RL-based baselines. As part of this effort, we plan to open-source the digital twin and associated VLM–RL training stack as a benchmark environment for UAV wildfire perception, path planning, and monitoring tasks, with standardized scenarios, metrics, and ablation protocols.
Future work will extend this framework to cooperative multi-UAV planning and more realistic fire–atmosphere coupling, and explore wildfire-specific VLM adaptation using curated RGB/thermal datasets. We also aim to port the pipeline to onboard hardware and study sim-to-real transfer, enabling field deployment of semantically guided UAV policies for real-world wildfire response.

\section*{Acknowledgments}
This material is based upon work supported by the National Science Foundation under Grant Nos. CNS-2232048 and CNS-2204445, and by NASA under award No. 80NSSC23K1393.

\newpage

{\small
\bibliographystyle{ieeenat_fullname}
\bibliography{main}
}


\clearpage
\appendix

\section{Implementation Details}
\subsection{PPO Modeling \& Hyperparameters}
\label{sec:ppo_hyperparams}

One of the most challenging aspects of implementing a reinforcement learning agent is hyperparameter tuning, since these choices strongly influence stability, convergence speed, and the final behavior of the learned policy. In our work, we adopt the Proximal Policy Optimization (PPO) implementation from Stable-Baselines3~\cite{sb3ppo} and treat the hyperparameters as a carefully tuned configuration for reliable training in the wildfire digital twin.

Table~\ref{tab:hyperparameters} summarizes the main PPO hyperparameters used for all experiments, including optimization settings, rollout configuration, and training horizon.

\begin{table*}[t]
\centering
\footnotesize
\renewcommand{\arraystretch}{1.3}
\setlength{\tabcolsep}{3pt}
\begin{tabular}{|c|c|c|c|}
\hline
\textbf{Hyperparameter} & \textbf{Name} &
\textbf{Value} & \textbf{Description} \\
\hline

Learning Rate  &  \texttt{learning\_rate}  &  $3\times 10^{-4}$  &  Adam optimizer learning rate  \\
\hline
Steps per Batch  &  \texttt{n\_steps}  &  2000  &  Environment steps per policy update  \\
\hline
Minibatch Size  &  \texttt{batch\_size}  &  400  &  Minibatch size for gradient updates  \\
\hline
Number of Epochs  &  \texttt{n\_epochs}  &  20  &  PPO epochs per update  \\
\hline
Discount Factor $\gamma$  &  \texttt{gamma}  &  0.99  &  Return discount factor  \\
\hline
Loss Function  &  \texttt{loss\_func}  &  \texttt{ClipPPOLoss}  &  Clipped PPO surrogate loss  \\
\hline
Value Estimator  &  \texttt{value\_estimator}  &  \texttt{GAE}  &  Generalized advantage estimation  \\
\hline
GAE $\lambda$  &  \texttt{gae\_lambda}  &  0.95  &  Smoothing parameter for GAE  \\
\hline
Clip Epsilon $\epsilon$  &  \texttt{clip\_range}  &  0.20  &  PPO clipping range  \\
\hline
Entropy Coefficient  &  \texttt{ent\_coef}  &  0.01  &  Entropy regularization coefficient  \\
\hline
Value Function Coeff.  &  \texttt{vf\_coef}  &  0.6  &  Value function loss coefficient  \\
\hline
Max Gradient Norm  &  \texttt{max\_grad\_norm}  &  0.5  &  Gradient clipping norm  \\
\hline
Steps per Episode  &  \texttt{max\_steps}  &  4000  &  Maximum timesteps per episode  \\
\hline
Evaluation Frequency  &  \texttt{eval\_freq}  &  20000  &  Evaluation callback frequency  \\
\hline
Total Timesteps  &  \texttt{total\_timesteps}  &  200000  &  Total training timesteps  \\
\hline
\end{tabular}
\caption{PPO hyperparameters used for training the wildfire-monitoring UAV policy with Stable-Baselines3~\cite{sb3ppo}.}
\label{tab:hyperparameters}
\end{table*}

\section{Extended Quantitative Results}
\label{sec:appendix_results}

This appendix reports the full quantitative results for all six models across the five evaluation tasks described in Section~\ref{sec:results}. Each table summarizes performance for a single task in terms of total return, wildfire visibility, detection latency, distance traveled, and runtime. Together, these results highlight how VLM-guided reward shaping, dual-camera sensing, and directional guidance contribute to improved wildfire localization, tracking, and robustness.

\subsection{Task 1: Wildfire in Initial Field of View}
\label{app:task1}

In Task~1, wildfire is already visible in the initial top-down frame, so all models can detect and track the fire without significant exploration. As expected, all six variants achieve 100\% time-in-FOV and near-identical time-to-detection. Differences therefore arise mainly in total reward and flight efficiency. The VLM-guided final model attains the highest total reward while maintaining 100\% coverage, indicating slightly more efficient control and reward optimization, whereas the VLM-only model travels substantially less distance due to the absence of a base exploration reward.

\begin{table}[h]
\centering
\scriptsize
\renewcommand{\arraystretch}{1.3}
\setlength{\tabcolsep}{3pt}
\begin{tabular}{|c|c|c|c c|c|c|}
\hline
\textbf{Model} & \textbf{R$_{total}$} &
\textbf{FOV\%} & \textbf{TTD (s)} &
\textbf{(steps)} & \textbf{Dist (m)}
& \textbf{Runtime (s)} \\
\hline
\textbf{Base PPO}  &  5112.54  &  \textbf{100.0\%}  &  \textbf{6.12}  &  \textbf{1.0}  &  4119.91  &  \textbf{4420.2}  \\
\textbf{ }  &    &     &    &    &     &     \\
\hline
\textbf{VLM-only}  &  3936.43  &  \textbf{100.0\%}  &  6.33  &  \textbf{1.0}  &  2028.45  &  4510.1  \\
\textbf{rew. shaping}  &    &     &    &    &     &     \\
\hline
\textbf{VLM-int.,}  &  5118.06  &  \textbf{100.0\%}  &  6.38  &  \textbf{1.0}  &  \textbf{6089.79}  &  4708.9  \\
\textbf{top-down only}  &    &     &    &    &     &     \\
\hline
\textbf{VLM-int.,}  &  4612.58  &  \textbf{100.0\%}  &  6.42  &  \textbf{1.0}  &  6076.34  &  4710.8  \\
\textbf{angled only}  &    &     &    &    &     &     \\
\hline
\textbf{VLM-guided,}  &  5129.60  &  \textbf{100.0\%}  &  6.33  &  \textbf{1.0}  &  6078.10  &  4705.0  \\
\textbf{unsegmented}  &    &     &    &    &     &     \\
\hline
\textbf{VLM-guided}  &  \textbf{5146.59}  &  \textbf{100.0\%}  &  6.32  &  \textbf{1.0}  &  6073.74  &  4707.1  \\
\textbf{final model}  &    &     &    &    &     &     \\
\hline
\end{tabular}
\caption{Task 1 results: wildfire initially in the UAV's top-down FOV. All models achieve 100\% visibility; differences arise mainly in total reward and path characteristics.}
\label{tab:app_ablation1}
\end{table}

\subsection{Task 2: Wildfire Near UAV (Out of Initial FOV)}
\label{app:task2}

Task~2 requires short-range search: the UAV starts roughly 100\,m from the fire with no flames in the initial top-down view. Here, model differences become more pronounced. The VLM-guided final model achieves the highest total reward, highest time-in-FOV, and shortest time-to-detection, indicating more efficient exploration and transition from search to tracking. The VLM-guided model without segmentation performs similarly, while the baseline PPO and VLM-only variants show lower FOV coverage and much slower detection.

\begin{table}[h]
\centering
\scriptsize
\renewcommand{\arraystretch}{1.3}
\setlength{\tabcolsep}{3pt}
\begin{tabular}{|c|c|c|c c|c|c|}
\hline
\textbf{Model} & \textbf{R$_{total}$} &
\textbf{FOV\%} & \textbf{TTD (s)} &
\textbf{(steps)} & \textbf{Dist (m)}
& \textbf{Runtime (s)} \\
\hline
\textbf{Base PPO}  &  4278.30  &  84.8\%  &  669.23  &  596  &  3890.03  &  \textbf{4491.5}  \\
\textbf{ }  &    &     &    &    &     &     \\
\hline
\textbf{VLM-only}  &  3520.37  &  89.0\%  &  494.21  &  432  &  2048.45  &  4576.2  \\
\textbf{rew. shaping}  &    &     &    &    &     &     \\
\hline
\textbf{VLM-int.,}  &  4674.29  &  91.3\%  &  332.52  &  280  &  5218.33  &  4697.3  \\
\textbf{top-down only}  &    &     &    &    &     &     \\
\hline
\textbf{VLM-int.,}  &  3910.21  &  88.2\%  &  375.28  &  290  &  5436.07  &  4706.7  \\
\textbf{angled only}  &    &     &    &    &     &     \\
\hline
\textbf{VLM-guided,}  &  5001.84  &  96.5\%  &  182.69  &  116  &  \textbf{5721.53}  &  4708.1  \\
\textbf{unsegmented}  &    &     &    &    &     &     \\
\hline
\textbf{VLM-guided}  &  \textbf{5018.22}  &  \textbf{97.1\%}  &  \textbf{180.05}  &  \textbf{113}  &  5710.59  &  4716.0  \\
\textbf{final model}  &    &     &    &    &     &     \\
\hline
\end{tabular}
\caption{Task 2 results: wildfire near the UAV ($\approx 100$\,m) but initially outside the top-down FOV. The VLM-guided models detect the fire substantially faster and achieve higher FOV coverage.}
\label{tab:app_ablation2}
\end{table}

\subsection{Task 3: Wildfire at Long Range}
\label{app:task3}

Task~3 is the most challenging setting: the UAV starts roughly 1\,km from the fire, with no flames in the top-down view and only a small smoke cue in the angled view. Under this long-range, sparse-reward regime, only the two dual-camera VLM-guided models succeed in detecting the wildfire within the 4000-timestep limit. All four other models fail to bring the fire into the top-down FOV, yielding zero FOV time and undefined detection latency. The final model achieves the highest total reward, non-zero FOV coverage, and a finite time-to-detection, highlighting the importance of VLM-based semantic guidance for distant search.

\begin{figure}[t]
\centering
\includegraphics[width=.8\linewidth]{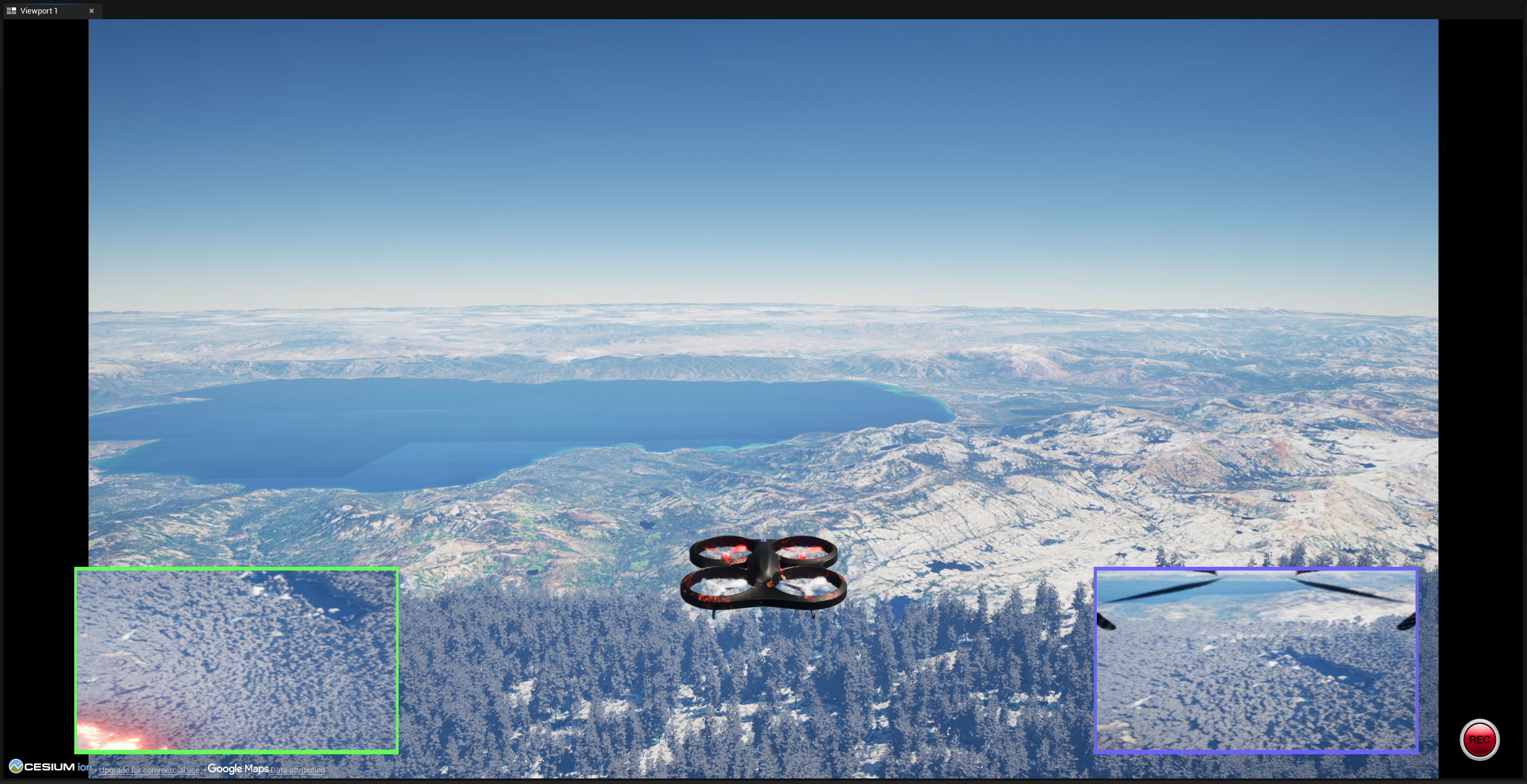}
\caption{UAV deployed with wildfire in the initial FOV of its top-down camera (Wildfire in FOV).}
\label{fig:task1deployment}
\end{figure}

\begin{figure}[t]
\centering
\includegraphics[width=.8\linewidth]{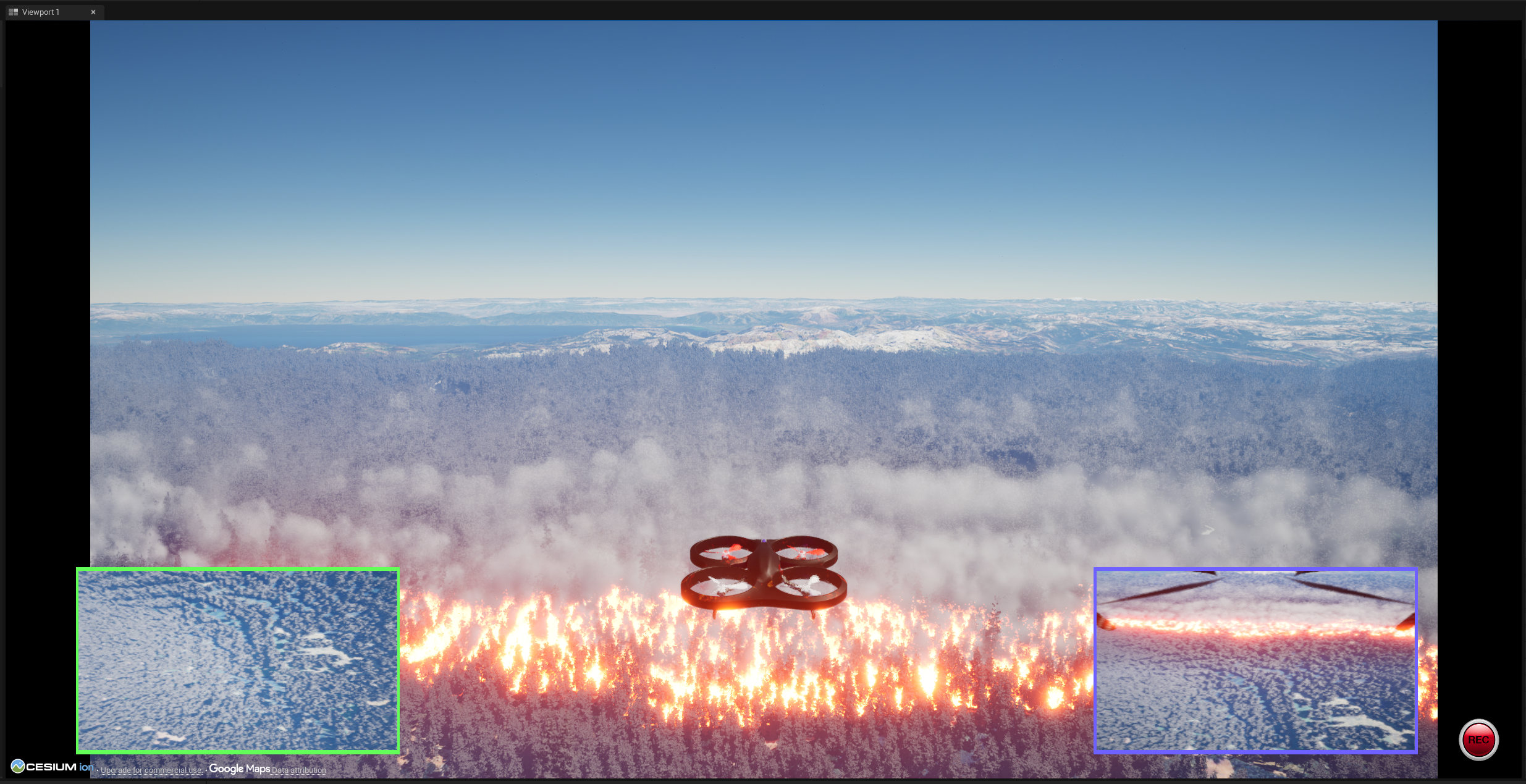}
\caption{UAV deployed approximately 100 m away from the nearest wildfire instance (Wildfire near UAV).}
\label{fig:task2deployment}
\end{figure}

\begin{figure}[t]
\centering
\includegraphics[width=.8\linewidth]{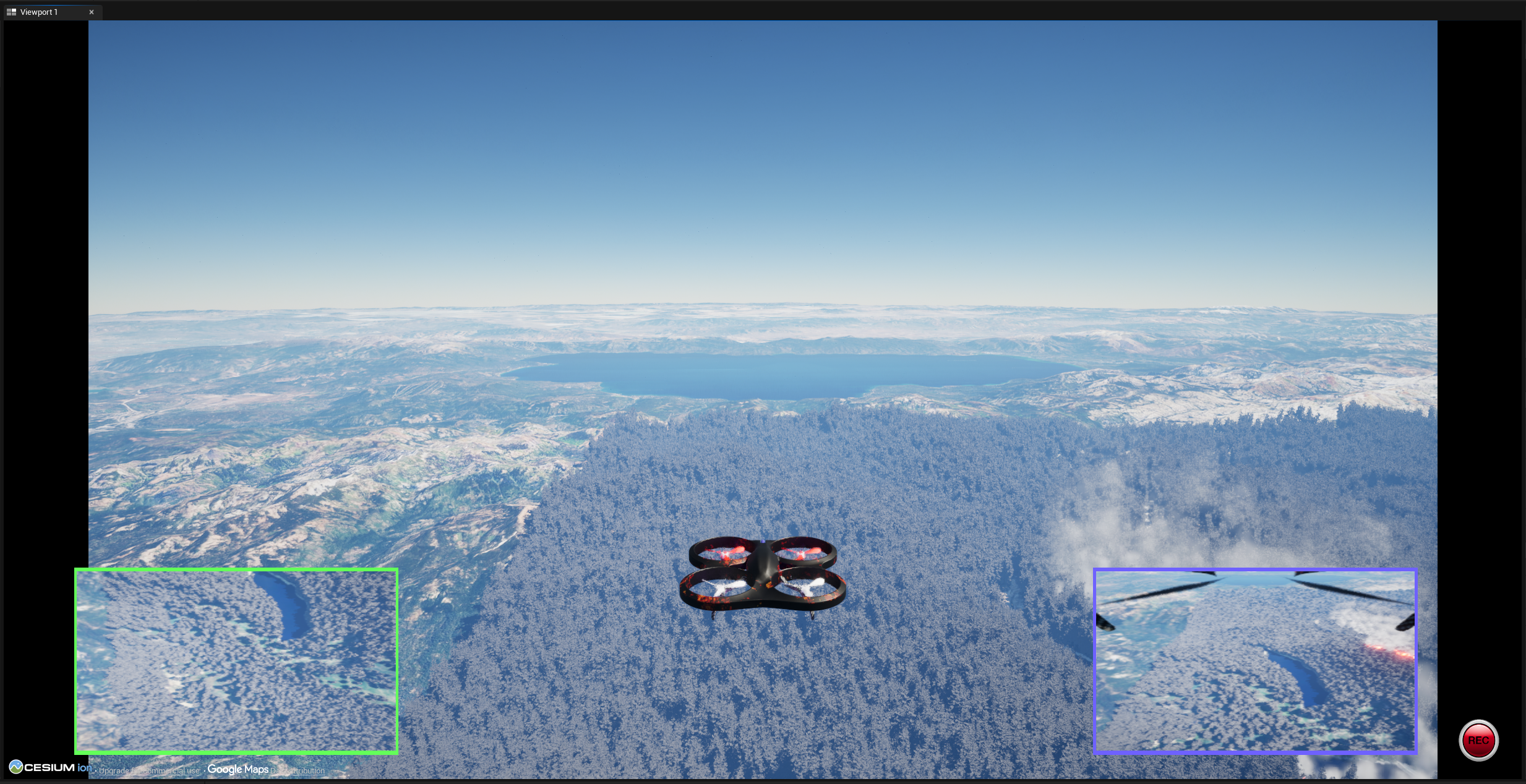}
\caption{UAV deployed approximately 1 km away from the nearest wildfire instance (Wildfire in distance).}
\label{fig:task3deployment}
\end{figure}

\begin{table}[h]
\centering
\scriptsize
\renewcommand{\arraystretch}{1.3}
\setlength{\tabcolsep}{3pt}
\begin{tabular}{|c|c|c|c c|c|c|}
\hline
\textbf{Model} & \textbf{R$_{total}$} &
\textbf{FOV\%} & \textbf{TTD (s)} &
\textbf{(steps)} & \textbf{Dist (m)}
& \textbf{Runtime (s)} \\
\hline
\textbf{Base PPO}  &  486.39  &  0.0\%  &  None  &  None  &  3513.58  &  \textbf{4238.4}  \\
\textbf{ }  &    &     &    &    &     &     \\
\hline
\textbf{VLM-only}  &  314.82  &  0.0\%  &  None  &  None  &  1596.23  &  4378.5  \\
\textbf{rew. shaping}  &    &     &    &    &     &     \\
\hline
\textbf{VLM-int.,}  &  629.13  &  0.0\%  &  None  &  None  &  1902.76  &  4511.7  \\
\textbf{top-down only}  &    &     &    &    &     &     \\
\hline
\textbf{VLM-int.,}  &  637.54  &  0.0\%  &  None  &  None  &  2562.39  &  4496.5  \\
\textbf{angled only}  &    &     &    &    &     &     \\
\hline
\textbf{VLM-guided,}  &  2953.47  &  42.7\%  &  2684.19  &  2210  &  4484.91  &  4640.1  \\
\textbf{unsegmented}  &    &     &    &    &     &     \\
\hline
\textbf{VLM-guided}  &  \textbf{3612.46}  &  \textbf{45.1\%}  &  \textbf{2492.54}  &  \textbf{2118}  &  \textbf{4875.80}  &  4656.4  \\
\textbf{final model}  &    &     &    &    &     &     \\
\hline
\end{tabular}
\caption{Task 3 results: wildfire at long range ($\approx 1$\,km). Only the two dual-camera VLM-guided models successfully detect and track the distant fire within 4000 timesteps.}
\label{tab:app_ablation3}
\end{table}

\subsection{Task 4: Robustness to Wind}
\label{app:task4}

Task~4 reuses the Task~2 spatial configuration (wildfire $\approx 100$\,m away, initially outside the top-down FOV) but adds a 5\,m/s crosswind. All models experience some degradation relative to Task~2, yet the VLM-guided final model again attains the highest total reward, high FOV coverage, and low detection latency. The close performance between the final model and the VLM-guided variant without segmentation suggests that dual-view VLM integration is the primary driver of robustness, with directional guidance providing an additional refinement.

\begin{table}[h]
\centering
\scriptsize
\renewcommand{\arraystretch}{1.3}
\setlength{\tabcolsep}{3pt}
\begin{tabular}{|c|c|c|c c|c|c|}
\hline
\textbf{Model} & \textbf{R$_{total}$} &
\textbf{FOV\%} & \textbf{TTD (s)} &
\textbf{(steps)} & \textbf{Dist (m)}
& \textbf{Runtime (s)} \\
\hline
\textbf{Base PPO}  &  4291.64  &  83.5\%  &  676.90  &  602  &  4236.58  &  \textbf{4486.1}  \\
\textbf{ }  &    &     &    &    &     &     \\
\hline
\textbf{VLM-only}  &  3538.61  &  86.4\%  &  492.76  &  429  &  2416.52  &  4572.3  \\
\textbf{rew. shaping}  &    &     &    &    &     &     \\
\hline
\textbf{VLM-int.,}  &  4690.39  &  88.9\%  &  358.29  &  292  &  5505.19  &  4689.4  \\
\textbf{top-down only}  &    &     &    &    &     &     \\
\hline
\textbf{VLM-int.,}  &  3891.02  &  86.5\%  &  388.17  &  294  &  5617.43  &  4691.3  \\
\textbf{angled only}  &    &     &    &    &     &     \\
\hline
\textbf{VLM-guided,}  &  5012.85  &  93.7\%  &  187.34  &  120  &  \textbf{6030.84}  &  4706.9  \\
\textbf{unsegmented}  &    &     &    &    &     &     \\
\hline
\textbf{VLM-guided}  &  \textbf{5034.62}  &  \textbf{96.9\%}  &  \textbf{183.61}  &  \textbf{115}  &  5925.30  &  4711.2  \\
\textbf{final model}  &    &     &    &    &     &     \\
\hline
\end{tabular}
\caption{Task 4 results: wildfire near the UAV with added crosswind ($\approx 5$\,m/s). The VLM-guided models remain robust under plume-induced drift and environmental disturbance.}
\label{tab:app_ablation4}
\end{table}

\subsection{Task 5: Robustness to Low Lighting}
\label{app:task5}

Task~5 again mirrors the Task~2 geometry but under low-light conditions to emulate nighttime or low-visibility operation. The VLM-guided final model achieves the highest total reward, highest FOV coverage, and fastest detection, indicating that VLM-based semantic cues help compensate for reduced RGB contrast. Notably, all VLM-integrated models outperform the baseline PPO in both time-in-FOV and time-to-detection, suggesting that wildfire-specific semantics (flame glow, plume structure) remain informative even when the overall scene is dark.

\begin{table}[h]
\centering
\scriptsize
\renewcommand{\arraystretch}{1.3}
\setlength{\tabcolsep}{3pt}
\begin{tabular}{|c|c|c|c c|c|c|}
\hline
\textbf{Model} & \textbf{R$_{total}$} &
\textbf{FOV\%} & \textbf{TTD (s)} &
\textbf{(steps)} & \textbf{Dist (m)}
& \textbf{Runtime (s)} \\
\hline
\textbf{Base PPO}  &  4291.65  &  85.0\%  &  601.84  &  595  &  3891.20  &  \textbf{4015.6}  \\
\textbf{ }  &    &     &    &    &     &     \\
\hline
\textbf{VLM-only}  &  3751.94  &  89.0\%  &  443.75  &  429  &  2052.96  &  4056.0  \\
\textbf{rew. shaping}  &    &     &    &    &     &     \\
\hline
\textbf{VLM-int.,}  &  4859.16  &  91.5\%  &  290.30  &  282  &  5216.71  &  4091.3  \\
\textbf{top-down only}  &    &     &    &    &     &     \\
\hline
\textbf{VLM-int.,}  &  3992.37  &  88.8\%  &  292.43  &  289  &  5432.15  &  4073.2  \\
\textbf{angled only}  &    &     &    &    &     &     \\
\hline
\textbf{VLM-guided,}  &  5015.07  &  97.0\%  &  117.90  &  112  &  \textbf{5710.34}  &  4102.6  \\
\textbf{unsegmented}  &    &     &    &    &     &     \\
\hline
\textbf{VLM-guided}  &  \textbf{5039.43}  &  \textbf{97.4\%}  &  \textbf{112.96}  &  \textbf{108}  &  5686.39  &  4084.5  \\
\textbf{final model}  &    &     &    &    &     &     \\
\hline
\end{tabular}
\caption{Task 5 results: wildfire near the UAV under low-light conditions. The VLM-guided models maintain high visibility and rapid detection despite reduced RGB contrast.}
\label{tab:app_ablation5}
\end{table}

\subsection{Qualitative Reward and Trajectory Trends}
\label{app:qualitative}

Beyond scalar metrics, these results show consistent qualitative patterns across Tasks~1–3. Reward curves exhibit three phases: an initial exploration phase, a stabilization phase as the agent converges toward the frontline, and a monitoring phase with relatively steady or increasing reward once the fire remains in view. Smoothed reward curves (50-step moving average) highlight this structure by reducing short-term variance.

Trajectory and altitude plots further confirm the effect of the shaped base reward: the agent typically ascends rapidly early in the episode, then slows its climb and emphasizes horizontal coverage as it begins tracking the fireline. Velocity traces show high variability during early exploration, followed by more stable speeds once the agent locks onto the firefront. These extended results collectively show that the final VLM-guided model not only achieves higher quantitative scores but also exhibits more structured, interpretable flight behavior across diverse wildfire scenarios.


\section{Reward Curves Across Tasks}
\label{app:reward_curves}

We also report full reward trajectories for the final VLM-guided model over the three core tasks (wildfire in FOV, wildfire near UAV, wildfire in distance). For each task we plot both the raw per-timestep reward and a 50-step moving average to visualize the underlying trend.

\begin{figure}[t]
\centering
\begin{minipage}{0.48\linewidth}
    \centering
    \includegraphics[width=\linewidth]{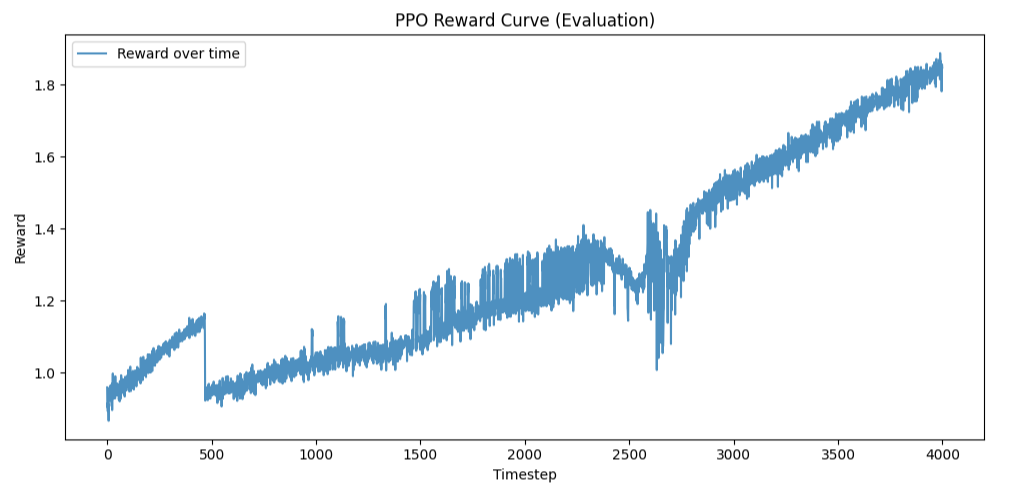}\\
    \scriptsize (a) Raw reward
\end{minipage}
\hfill
\begin{minipage}{0.48\linewidth}
    \centering
    \includegraphics[width=\linewidth]{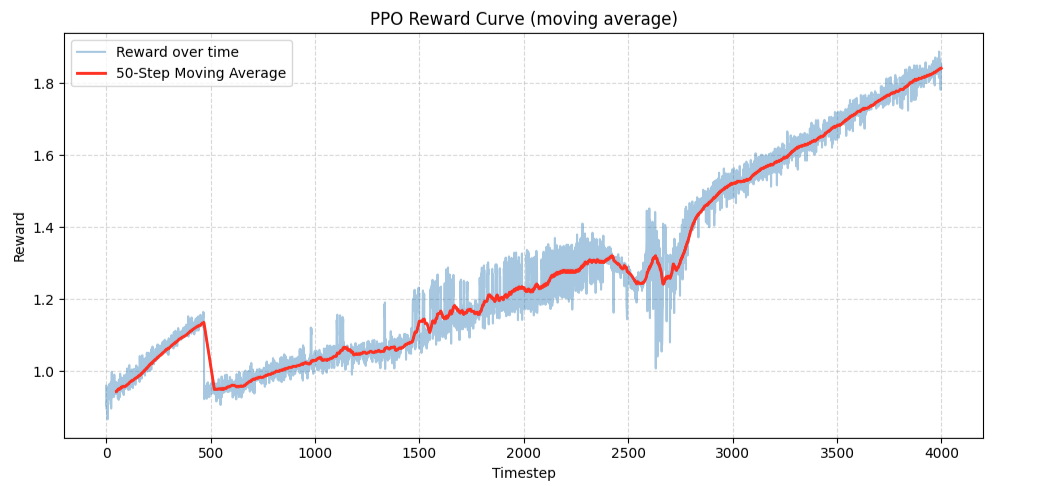}\\
    \scriptsize (b) 50-step moving average
\end{minipage}
\caption{Task~1 reward curves (wildfire in initial FOV).}
\label{fig:app_reward_task1}
\end{figure}

\begin{figure}[t]
\centering
\begin{minipage}{0.48\linewidth}
    \centering
    \includegraphics[width=\linewidth]{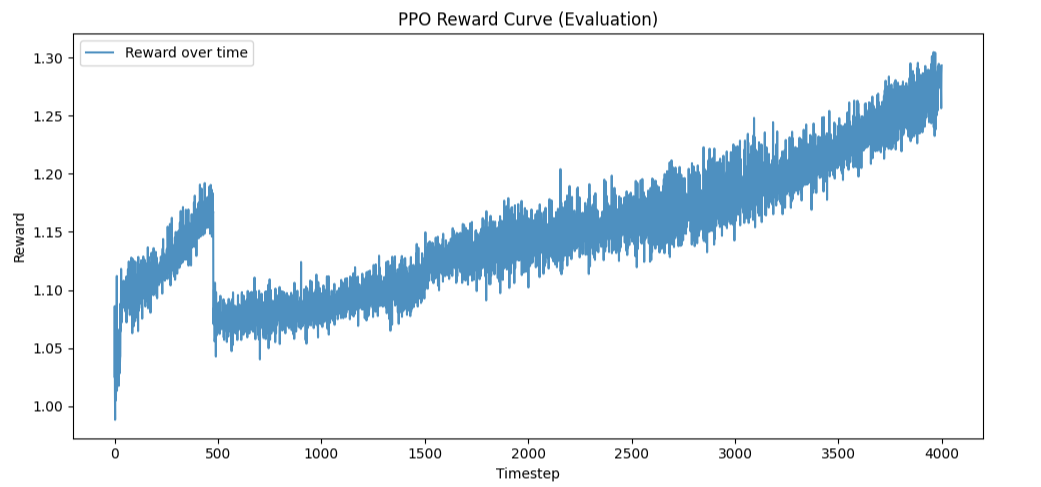}\\
    \scriptsize (a) Raw reward
\end{minipage}
\hfill
\begin{minipage}{0.48\linewidth}
    \centering
    \includegraphics[width=\linewidth]{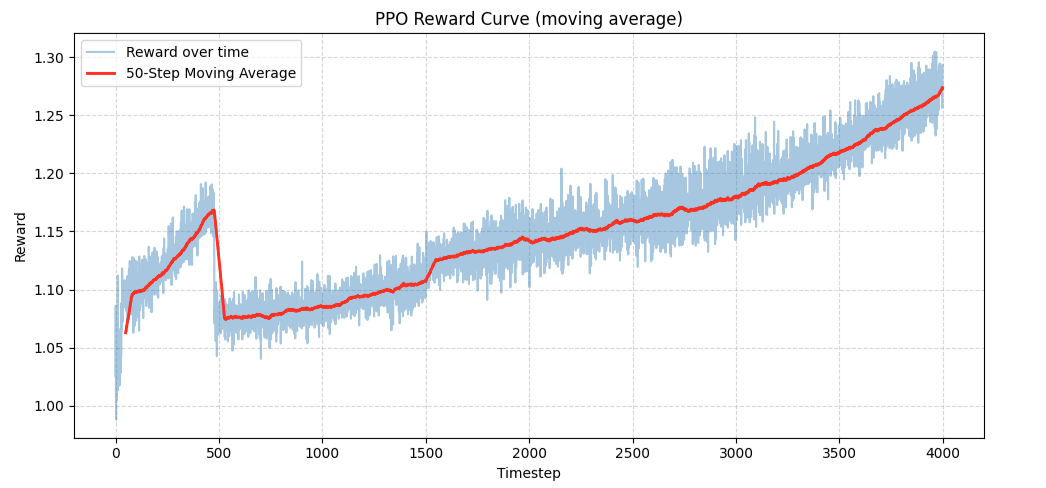}\\
    \scriptsize (b) 50-step moving average
\end{minipage}
\caption{Task~2 reward curves (wildfire near UAV, outside initial FOV).}
\label{fig:app_reward_task2}
\end{figure}

\begin{figure}[t]
\centering
\begin{minipage}{0.48\linewidth}
    \centering
    \includegraphics[width=\linewidth]{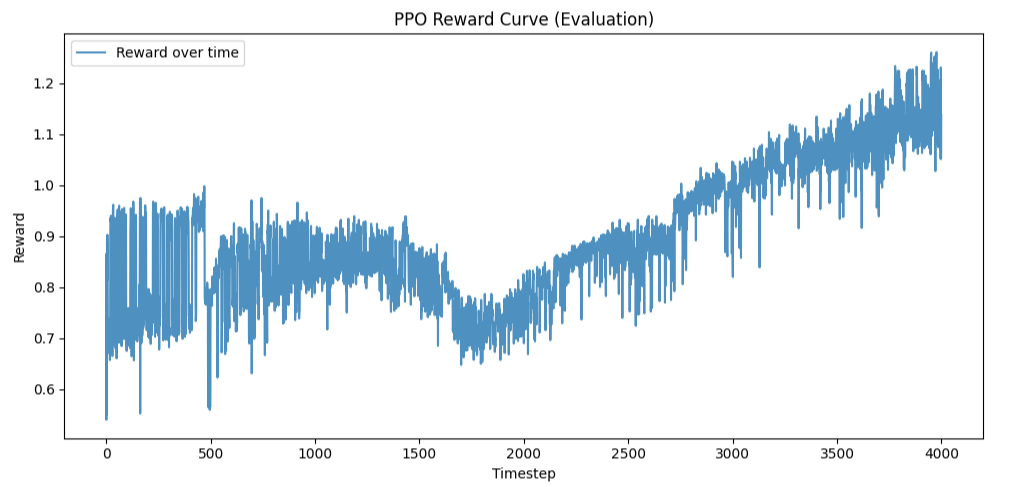}\\
    \scriptsize (a) Raw reward
\end{minipage}
\hfill
\begin{minipage}{0.48\linewidth}
    \centering
    \includegraphics[width=\linewidth]{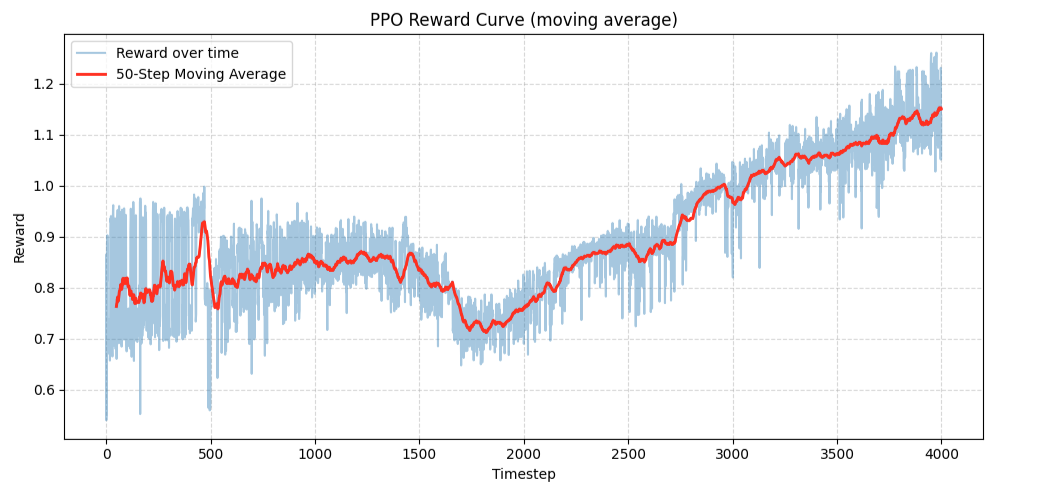}\\
    \scriptsize (b) 50-step moving average
\end{minipage}
\caption{Task~3 reward curves (wildfire at $\sim$1\,km distance).}
\label{fig:app_reward_task3}
\end{figure}

Across all three tasks, the smoothed curves show a consistent pattern: an initial high-variance exploration phase, followed by a stabilization phase as the agent converges toward the firefront, and a monitoring phase with relatively steady reward once the wildfire remains in view.

\end{document}